\documentclass{article}

\usepackage{arxiv}			
\usepackage[utf8]{inputenc} 
\usepackage[T1]{fontenc}    
\usepackage{url}            
\usepackage{booktabs}       

\usepackage{amsmath,amssymb,amsfonts,latexsym,amsthm,bm,breqn,dsfont,gensymb}
\usepackage{mathtools}
\usepackage{latexsym,breqn}
\usepackage{nicefrac}       
\usepackage{microtype}      
\usepackage{natbib}

\usepackage{graphicx,subfigure}
\usepackage{pgf}
\usepackage{pgffor}
\usepackage{tikz}

\usepackage{algorithm}
\usepackage[noend]{algpseudocode}
\makeatletter
\def\BState{\State\hskip-\ALG@thistlm}

\usepackage{ctable,multicol,multirow}
\usepackage{enumitem}

\usepackage{color}
\definecolor{green}{rgb}{0, 0.5, 0}
\usepackage[pdftex,
			bookmarks, bookmarksnumbered=true, bookmarksopen=true,
			colorlinks, citecolor=blue, filecolor=black, linkcolor=red, urlcolor=green,
			pdfauthor={López-Lopera, Andrés Felipe}, pdftitle={PhDEMSE},
			pdftoolbar=true, pdfmenubar=true, pdffitwindow = true
			]{hyperref}

\usepackage{pifont}
\definecolor{blue}{rgb}{0.12, 0.27, 0.71}
\definecolor{green}{rgb}{0, 0.5, 0}
\definecolor{orange}{rgb}{1,0.5,0.05}

\newcommand*\rotv{\rotatebox{90}}

\newcommand{\Bbeta}{\bm{\beta}}

\newcommand{\Bnu}{\bm{\nu}}

\newcommand{\Bell}{\bm{\ell}}
\newcommand{\Btheta}{\bm{\theta}} 
\newcommand{\Bphi}{\bm{\phi}} \newcommand{\BPhi}{\bm{\Phi}}
 \newcommand{\BPsi}{\bm{\Psi}}
 
 \newcommand{\BOmega}{\bm{\Omega}}

\newcommand{\Ba}{\bm{a}} \newcommand{\BA}{\textbf{A}}
\newcommand{\Bb}{\bm{b}} \newcommand{\BB}{\textbf{B}}
 
\newcommand{\BI}{\textbf{I}}

\newcommand{\Bf}{\bm{f}} \newcommand{\BF}{\bm{F}}

\newcommand{\Bu}{\bm{u}} \newcommand{\BU}{\textbf{U}}
\newcommand{\Bv}{\bm{v}} \newcommand{\BV}{\textbf{V}}
\newcommand{\Bk}{\bm{k}} \newcommand{\BK}{\textbf{K}}
\newcommand{\BL}{\textbf{L}}
\newcommand{\Bx}{\bm{x}} 
\newcommand{\By}{\bm{y}} \newcommand{\BY}{\textbf{Y}}

\newcommand{\Balpha}{\bm{\alpha}}
\newcommand{\realset}[1]{\mathbb{R}^{#1}}
\newcommand{\GP}[2]{\mathcal{GP}({#1,#2})}

\newcommand{\BfR}{\bm{\mathcal{R}}}

\newcommand{\BfF}{\bm{\mathcal{F}}}

\newcommand{\BfG}{\bm{\mathcal{G}}}

\newcommand{\swl}{\mbox{SWL}}
\newcommand{\msl}{\mbox{MSL}}
\newcommand{\tide}{\mbox{T}}
\newcommand{\surge}{\mbox{S}}
\newcommand{\Hs}{\mbox{Hs}}
\newcommand{\Tp}{\mbox{Tp}}
\newcommand{\Dp}{\mbox{Dp}}
\newcommand{\U}{\mbox{U}}
\newcommand{\Du}{\mbox{Du}}
\newcommand{\EFP}{\mbox{EFP}}

\RequirePackage[capitalize,nameinlink]{cleveref}[0.19]
\crefname{chapter}{chapter}{chapters}
\crefname{section}{section}{sections}
\crefname{subsection}{subsection}{subsections}

\Crefname{figure}{Figure}{Figures}
\Crefname{Condition}{Condition}{Condition}
\Crefname{Assumption}{Assumption}{Assumption}

\crefformat{equation}{\textup{#2(#1)#3}}
\crefrangeformat{equation}{\textup{#3(#1)#4--#5(#2)#6}}
\crefmultiformat{equation}{\textup{#2(#1)#3}}{ and \textup{#2(#1)#3}}
{, \textup{#2(#1)#3}}{, and \textup{#2(#1)#3}}
\crefrangemultiformat{equation}{\textup{#3(#1)#4--#5(#2)#6}}%
{ and \textup{#3(#1)#4--#5(#2)#6}}{, \textup{#3(#1)#4--#5(#2)#6}}{, and \textup{#3(#1)#4--#5(#2)#6}}

\Crefformat{equation}{#2Equation~\textup{(#1)}#3}
\Crefrangeformat{equation}{Equations~\textup{#3(#1)#4--#5(#2)#6}}
\Crefmultiformat{equation}{Equations~\textup{#2(#1)#3}}{ and \textup{#2(#1)#3}}
{, \textup{#2(#1)#3}}{, and \textup{#2(#1)#3}}
\Crefrangemultiformat{equation}{Equations~\textup{#3(#1)#4--#5(#2)#6}}%
{ and \textup{#3(#1)#4--#5(#2)#6}}{, \textup{#3(#1)#4--#5(#2)#6}}{, and \textup{#3(#1)#4--#5(#2)#6}}

\title{Multioutput Gaussian Processes with Functional \\ Data: A Study on Coastal Flood Hazard Assessment}
\date{}
\author{
  Andr\'es F. L\'opez-Lopera$^{\ast}$ \\ 
  {\fontsize{10}{10} \selectfont andres.lopezlopera@uphf.fr}
  \And
   D\'eborah Idier$^\ddagger$\\
  {\fontsize{10}{10} \selectfont d.idier@brgm.fr}  
  \And
  J\'er\'emy Rohmer$^\ddagger$\\
  {\fontsize{10}{10} \selectfont j.rohmer@brgm.fr}   
  \AND
  Fran\c{c}ois Bachoc$^\dagger$\\
  {\fontsize{10}{10} \selectfont francois.bachoc@math.univ-toulouse.fr}
  \AND
  \vspace{-3ex}
  {} \\ 
  $^\ast$ Universit\'e Polytechnique Hauts-de-France, C\'eramaths, F-59313 Valenciennes, France\\
  $^\dagger$ Institut de Math\'ematiques de Toulouse, Universit\'e Paul Sabatier, F-31062 Toulouse, France \\
  $^\ddagger$ BRGM, DRP/R3C, 3 avenue Claude Guillemin, F-45060 Orl\'eans c\'edex 2, France
}

\begin{document}
\maketitle

\vskip-7ex
\begin{abstract}
	Surrogate models are often used to replace costly-to-evaluate complex coastal codes to achieve substantial computational savings. In many of those models, the hydrometeorological forcing conditions (inputs) or flood events (outputs) are conveniently parameterized by scalar representations, neglecting that the inputs are actually time series and that floods propagate spatially inland. Both facts are crucial in flood prediction for complex coastal systems. Our aim is to establish a surrogate model that accounts for time-varying inputs and provides information on spatially varying inland flooding. We introduce a multioutput Gaussian process model based on a separable kernel that correlates both functional inputs and spatial locations. Efficient implementations consider tensor-structured computations or sparse-variational approximations. In several experiments, we demonstrate the versatility of the model for both learning maps and inferring unobserved maps, numerically showing the convergence of predictions as the number of learning maps increases. We assess our framework in a coastal flood prediction application. Predictions are obtained with small error values within computation time highly compatible with short-term forecast requirements (on the order of minutes compared to the days required by hydrodynamic simulators). We conclude that our framework is a promising approach for forecast and early-warning systems.
\end{abstract}


\section{Introduction}
\label{sec:intro}

Natural hazards, such as floods induced by Hurricane Katrina (2005) or by the more recent storms Xynthia (2010) and Johanna (2008), have strong negative impacts on the living conditions of hundreds of people \citep{Blake2016TechReportCyclones,Lumbroso2011CoastalFlooding,Andre2013CoastalFlooding,BATHRELLOS2017119}. These events caused harmful coastal flooding that resulted in a significant number of deaths. Hurricane Katrina was one of the six most powerful hurricanes ever recorded in the Atlantic. It resulted in a death toll of 1836 and approximately 80 billion dollars of damage \citep{Blake2016TechReportCyclones}. 
The storm Xynthia severely impacted low-lying French coastal areas located in the central part of the Bay of Biscay on February 27 and 28, 2010 \citep{Bertin2012Xynthia}. The flood induced by Xynthia caused 53 fatalities and more than 1 billion euros of material damage. The storm Johanna had minor effects on the French Atlantic coast but still led to significant flooding damage, such as in the town of G\^avres \citep[Britany; ][]{Andre2013CoastalFlooding,Idier2020CoastalFA}. These historical flood episodes reflect the need for accurate forecast and early-warning systems (FEWSs) to reduce the loss of human life and damage in areas at risk of flooding \citep{Andre2013CoastalFlooding,Hoggart2014CoastalFloodingSurvey,Idier2020CoastalFA}.

Most of the existing coastal flood FEWSs do not model floods and rely on the prediction of hydrodynamic conditions on the coast and expert judgment \citep[see, e.g.,][]{Doong2012EarlyWarningSystem}. Some systems under development rely on simplified inland flood computations \citep[see, e.g.,][]{Tromble2019HydroModelling,Stansby2013CoastalFlooding}, expecting that high-performance computing will allow their integration in operational platforms. In the Netherlands, the operational FEWS issues warnings based on the forecast of the near-shore conditions (like many other FEWSs), but in the case of warnings, floods are estimated using a database of precomputed flooding scenarios. These scenarios were generated for a limited set of dike breach locations and water levels using a model resolution of approximately 10 m. Most of the recent contributions allow modeling of high-resolution floods, even if wave overtopping plays an important role \citep[see, e.g.,][]{LeRoy2015Johanna,Idier2020CoastalFA}. However, those models are costly to evaluate, requiring up to days of parallel computations, making their use in warning forecasting systems impractical.

To overcome the computational complexity of coastal flooding codes, data-driven surrogate models have been widely explored \citep{Sacks1989CompExps,Rohmer2016CycloneSensitivity,Liu2017_GPTsunami,Li2020Surrogate,Bolle2018BayesianDecisionSupport,Rueda2019HyCReWW}. These models are initially fed with a statistically rich but tractable number of simulations of a former model. By learning statistical features, surrogate models are then used to predict floods based on knowledge of the coastal state. As shown in \citep{Rohmer2012CoastalFlooding,Jia2013GPhurricane,Liu2017_GPTsunami,Azzimonti2018CoastalFlooding,Betancourt2020fGPs,Perrin2020fPCAGP}, Gaussian process (GP)-based stochastic surrogates can be successfully applied in a wide range of coastal engineering applications. {More precisely, they are often used as surrogate models for expensive codes that can be later exploited, such as in regression and sensitivity analysis, both of which are of great interest for reliability engineering and system safety \citep{Betancourt2020fGPs,Perrin2020fPCAGP,Quoc2021GPRelicitedData,DaVeiga2020constrGP,Li2020Surrogate,Iooss2009SensitivityAnalysis,Marrel2009SensitivityGPs}.} In \citep{Rohmer2012CoastalFlooding,Azzimonti2018CoastalFlooding}, GP-based approaches are used to assess the impact of critical spatial offshore conditions. Both frameworks address level set estimation problems aiming to identify when the offshore conditions exceed a fixed risk threshold. In \citep{Jia2013GPhurricane,Liu2017_GPTsunami}, the authors focused on the dimension reduction of expensive computer codes via GP emulators for tsunamis and storm/hurricane risk assessment, respectively. {In \citep{Li2020Surrogate}, a GP-based sensitivity analysis study in the context of expensive models with high-dimensional outputs is conducted. Additional works related to dimension reduction and GPs have been proposed in \citep{Fukutani2021PCAKriging,Rohmer2016CycloneSensitivity,Perrin2020fPCAGP}}

Many of the aforementioned works have in common that they consider scalar representations of the hydrodynamic forcing conditions (inputs) rather than their functional structures (e.g., time series). We particularly refer to \citep{Azzimonti2018CoastalFlooding,Rohmer2018RandomForest} for some examples where hydrodynamic functional drivers are parameterized by scalar representations before assessing the surrogate models. To date, the frameworks in \citep{Betancourt2020fGPs,Kim2015funInputsHurricaine} are the only frameworks that consider time-varying inputs in the domains of flood hazard assessment and storm surge prediction, respectively. \citet{Betancourt2020fGPs} further investigate dedicated kernels based on proper distances in function spaces aiming to correlate hydrodynamic functional drivers. \citet{Kim2015funInputsHurricaine} introduce a time-dependent surrogate model of storm surge based on an artificial neural network.

\citet{Betancourt2020fGPs} have shown that considering hydrodynamic drivers as functions rather than scalars results in significant improvements in prediction as more precise physical information is encoded into kernels. Their work has focused on modeling, such as a scalar representation of the maximum cumulative overtopped and overflowed water volume. However, in practice, these global scalar indicators may be insufficient; and spatial information, e.g., the maximum inland water level, is often needed. {Therefore, inspired by \citep{Betancourt2020fGPs}, our aim here is to establish a GP model that accounts for the effects of time-varying forcing conditions and provides spatially varying inland flood information.} Our framework builds on the construction of a separable kernel that incorporates both functional and spatial correlations. The resulting process can be viewed as a multioutput GP \citep[see, e.g.][]{Alvarez2012kernelReview} where spatial flood indicators (outputs) are driven by a set of hydrometeorological drivers (functional inputs). In our case, the outputs are correlated by a kernel that exploits the ``similarity'' between the functional inputs. This leads to a framework that can be easily connected to other GP-based approaches \citep{Lazaro2009InterdomainGPs}. Furthermore, efficient implementations can be developed based on Kronecker-structured computations \citep[see, e.g.,][]{Alvarez2012kernelReview} or sparse-variational approximations \citep[see, e.g.,][]{GPflow2020multioutput}. For the former case, we provide R codes based on the \texttt{kergp} package \citep{Deville2019kergp}; and for the latter case, we adapt the multioutput GP models from the \texttt{GPflow} library \citep{GPflow2017} to account for functional input data.

There are alternative approaches that treat spatial outputs as functional data. For instance, we refer to \citep{Marrel2011GlobalSensitivitySpatGPs} for a framework modeling the pollution produced by radioactive wastes, to \citep{Chang2019ComputerCalibrationSpatialClimate} for an approach capable of learning spatial patterns in climate experiments, and to \citep{Perrin2020fPCAGP} for a surrogate model for the assessment of coastal flooding risks. These approaches first project the outputs into truncated basis representations (e.g., wavelets) to reduce the dimensionality. Then, prior distributions are placed at the level of the representation coefficients and not in the output space. Although their works scale well with the number of spatial points, they typically require a large number of learning simulations (e.g., over 500 events) to properly capture spatial patterns \citep[see, e.g.,][]{Perrin2020fPCAGP}. Here, we are restricted to highly constraining situations where fewer than 200 flood scenarios are available. This limitation arises from the use of costly-to-evaluate numerical simulators: a flood event requires almost three days of parallel computing \citep[see, e.g., the model used by][]{Idier2020CoastalFA}. We must also point out that \citep{Marrel2011GlobalSensitivitySpatGPs,Chang2019ComputerCalibrationSpatialClimate,Perrin2020fPCAGP} do not account for functional inputs, a condition that is crucial for properly learning hydrometeorological forcing conditions, as our framework does (see the discussion in Section \ref{sec:BRGMapp}). To the best of our knowledge, our proposal is the first to consider both inputs and outputs as functions in the field of flood hazard assessment. Furthermore, unlike the aforementioned works, our framework allows us to focus predictions on spatial design points placed in key sectors identified as of uttermost importance regarding the vulnerability of the territory \citep[see, e.g.,][for a further discussion]{Idier2013InverseMethodCoastalFlooding,Idier2020RISCOPE}.

The remaining sections are organized as follows. In Section \ref{sec:methods}, we describe the coastal flooding application that motivated the contributions in this paper. We briefly explain how to establish GP models for functional data, and we introduce the extension to spatial GPs with functional data where the covariance function is built via separable kernels. We also discuss the connection of the proposed framework with multioutput GP models. In Section \ref{sec:results}, we assess the performance of the resulting GP model on various synthetic examples considering different situations depending on the data availability. We also apply our framework to a coastal flooding application. Finally, in Section \ref{sec:conclusions}, we summarize our results and outline potential future work.


\section{Study Site, Data and Methods}
\label{sec:methods}

{In this section, we first describe the study site and the dataset used in the numerical experiments of Section \ref{sec:BRGMresults}. We then provide the background related to Gaussian process models with time-varying functional inputs. Finally, we introduce the extension to the case of spatial outputs.}

\subsection{Coastal flooding application: study site and dataset}
\label{sec:BRGMapp}
\begin{figure}[t!]
	\centering
	{\includegraphics[width = 0.65\textwidth,height=0.33\textwidth]{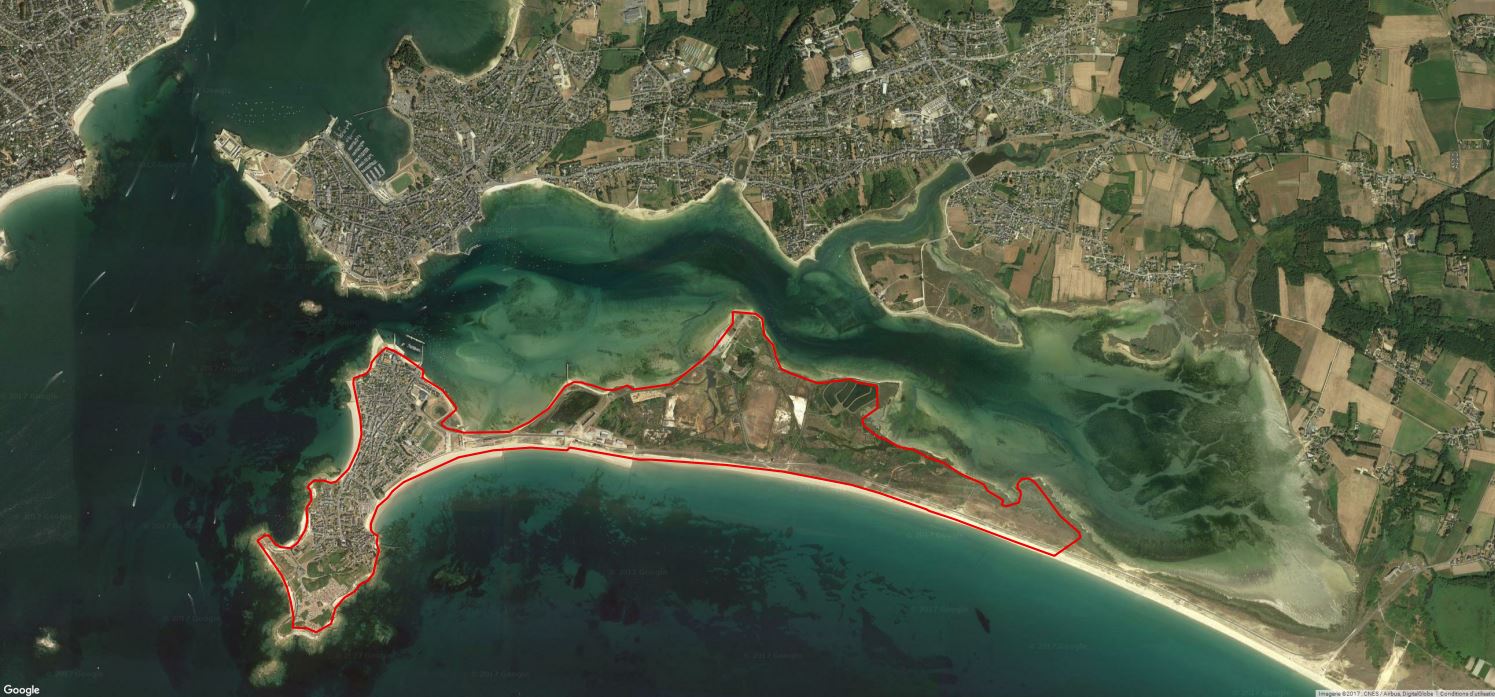}}
	\hskip0.5ex
	{\includegraphics[width = 0.27\textwidth,height=0.33\textwidth]{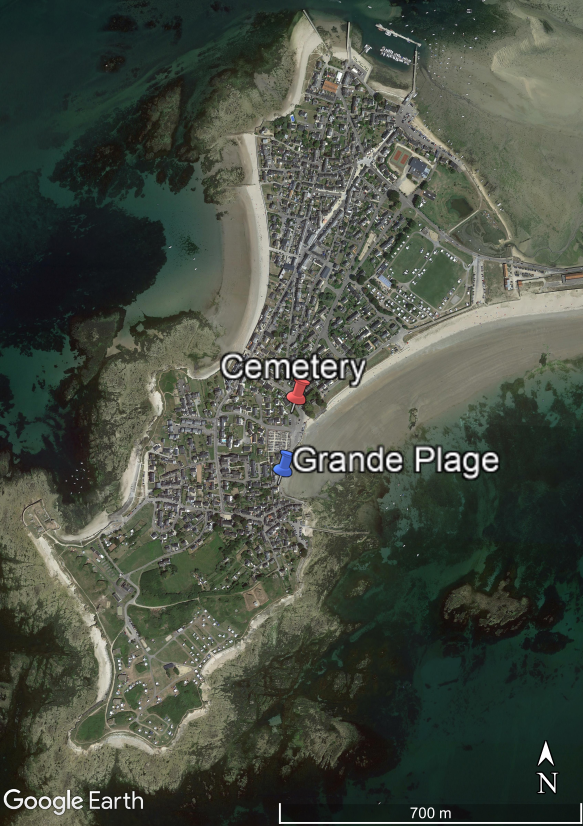}}\\
	{\footnotesize Map Data: Google, Maxar Technologies.}
	\caption{(left) The G\^avres municipality (red contour). (right) The town of G\^avres (\textit{Grande Plage} and the cemetery).}
	\label{fig:Gavres}
\end{figure}

In our study, we focus on the town of G\^avres (Figure \ref{fig:Gavres}, left) located along the French Atlantic coast in Brittany. The G\^avres municipality is a peninsula that is connected to the mainland by a 6 km long tombolo (Figure \ref{fig:Gavres}, right). The town has faced five significant coastal flood events since 1900 \citep{Idier2020CoastalFA}. The latest memorable event occurred on 10 March 2008 (Johanna storm): a combination of a spring tide, a storm surge larger than 0.5 m, and energetic waves led to the flooding of approximately 120 houses \citep{Idier2020CoastalFA}, some by approximately 1 m of water, along the street of the sports park \citep{LeRoy2015Johanna,Andre2013CoastalFlooding}. This marine submersion was induced mainly by waves overtopping the sea dike at the \textit{Grande-Plage} beach and a bit of overflow close to the cemetery (Figure \ref{fig:Gavres}, right).

In such environments, estimating floods requires the use of advanced hydrodynamic numerical models able to account for overflow and overtopping processes with good precision. Such models emulate the hydrodynamics (water level and current) induced by hydrometeorological forcing conditions (e.g., the mean sea level, tide, atmospheric surge, and wave conditions). An example is the nonhydrostatic phase-resolving SWASH model \citep{Zijlema2011SWASH}. The SWASH model, nested with a spectral wave model that propagates the offshore wave conditions to the SWASH model's boundary, can successfully reproduce local past flood events \citep[see ][]{Idier2020CoastalFA}. However, this modeling configuration is time consuming (6 h simulated in 3 days on 48 cores) and is therefore inapplicable for flood forecasting.
\begin{figure}[t!]
	\centering
	\foreach \i in {1,43,100} {
		\begin{minipage}{0.32\linewidth}
			\begin{tikzpicture}
				\node[anchor=south west,
				xshift=-\textwidth,
				yshift=-\textwidth] (image) at (current page.south west) {
					\includegraphics[width=\textwidth]{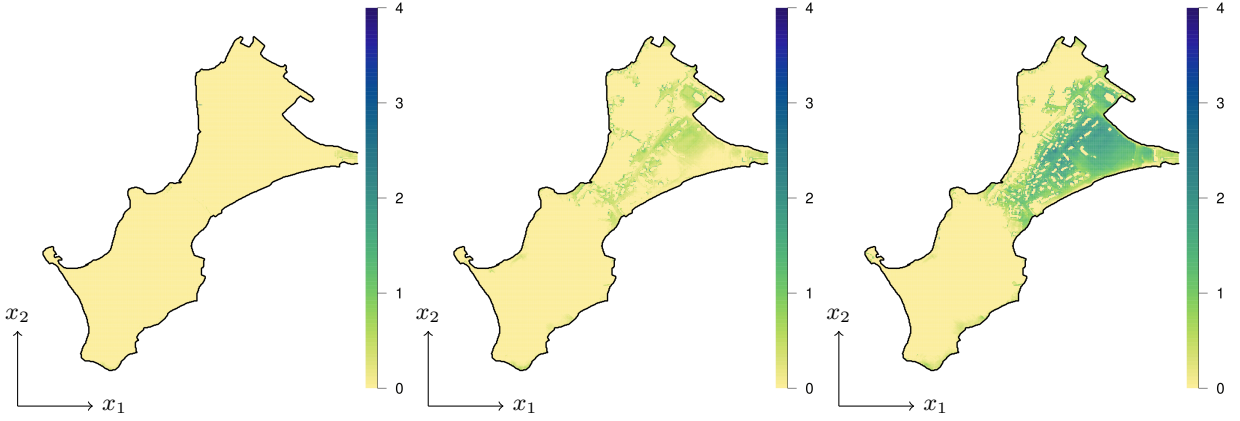}
				};
				\draw[->] (-\textwidth,-\textwidth)--++(0:1) node[right]{\footnotesize$x_1$};
				\draw[->] (-\textwidth,-\textwidth)--++(90:1) node[above]{\footnotesize$x_2$};
			\end{tikzpicture}
		\end{minipage}	
	}
	\caption{Three flood examples from the dataset in Section \ref{sec:BRGMapp}. The panels show the maximal inland water level $H_{\max}$ [m].}
	\label{fig:toyExample8BRGMMaps}		
\end{figure}

\begin{figure}[t!]
	\centering
	\foreach \i in {Slr,Tide,Surge,Tp,Hs,Dp,U,Du} {
		\includegraphics[width=0.24\linewidth]{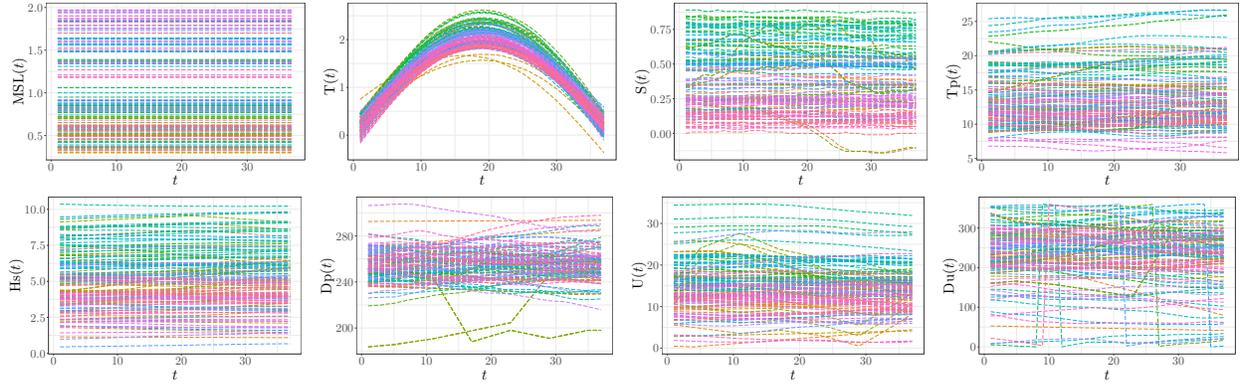}
	}
	\caption{Hydrometeorological functional inputs from the dataset in Section \ref{sec:BRGMapp}. The panels show the 131 replicates of the mean sea level ($\msl$ [m]), tide ($\tide$ [m]), surge ($\surge$ [m]), wave peak period ($\Tp$ [s]), wave height ($\Hs$ [m]), wave peak direction ($\Dp$ [$^\circ$]), wind speed ($\U$ [m/s]) and wind direction ($\Du$ [$^\circ$]). 
	}
	\label{fig:toyExample8BRGMInputs}
\end{figure}

Thus, to support the development of fast-running surrogate models for flood forecasting, the BRGM and IMT\footnote{BRGM: The French Geological Survey (acronym in French). IMT: Toulouse Mathematics Institute (acronym in French).} built a dataset $(X_o \to Y_o)$ based on numerical modeling. In the present work, we use this dataset. The hydrometeorological forcing conditions ($X_o$) are time series of the mean sea level ($\msl$ [m]), tide ($\tide$ [m]), atmospheric storm surge ($\surge$ [m]), significant wave height ($\Hs$ [m]), wave peak period ($\Tp$ [s]), wave peak direction ($\Dp$ [$^\circ$]), wind speed ($\U$ [m/s]) and wind direction ($\Du$ [$^\circ$]) \citep[see][for more details]{Idier2020CoastalFA}. These drivers, discretized with a 10 min time step over a 6 h window centered on high tide, are represented by a 37 observation long time series. 
The stored model results ($Y_o$) are the maximal inland water height ($H_{\max}$) at each grid point every 3 m (see Figure \ref{fig:toyExample8BRGMMaps}). This leads to spatial flood events containing approximately 64.6k inland observations. The dataset contains 131 scenarios of $X_o$, including 21 historical flood (9) and no flood (12) events \citep[see][]{Idier2020CoastalFA}; 16 scenarios simulated from small variations of the 9 historical flood events; and 94 additional scenarios with both zero, moderate and significant marine submersions. The latter (94) scenarios were built by applying a combination of methods to a hydrometeorological dataset covering the 1900-2016 period with a 10 min time step. Namely, multivariate extreme value analysis was applied to randomly generate the joint distribution of the maximum values of forcing conditions, a probabilistic classifier was applied to locate the time instant of these maximum values, and a multivariate Gaussian Monte Carlo-based sampling procedure was applied to generate the time series accordingly. We refer to \citep{Idier2020CoastalFA} for the dataset description and \citep{Rohmer2012CoastalFlooding,Idier2020CoastalFA} for the extreme value analysis. The 131 scenarios of the hydrometeorological functional inputs are shown in Figure \ref{fig:toyExample8BRGMInputs}.

\subsection{Background on Gaussian processes with functional inputs}
\label{sec:functionalGPs}

\subsubsection{Gaussian processes}
\label{sec:functionalGPs:subsec:GPs}
GP-based surrogate models have been widely used to replace costly-to-evaluate numerical simulators due to their well-founded and nonparametric framework for statistical learning \citep{Rasmussen2005GP,Camps2016GPSurvey}. GPs form a flexible prior over functions where regularity assumptions can be encoded into covariance functions also known as kernels \citep{Genton2001Kernels,Paciorek2004NonStCov}.

Let $\{Y(\BfF);  \BfF \in \mathcal{F}(\mathcal{T},\realset{})^Q \}$, where $\mathcal{F}(\mathcal{T},\realset{})$ is the set of functions from $\mathcal{T}$ to $\realset{}$, be a stochastic process with functional inputs $\BfF = (f_1, \ldots, f_Q)$. {Consistent with Section \ref{sec:BRGMapp}, here, we consider time-varying inputs $f : \mathcal{T} \to \realset{}$ with $\mathcal{T} \subseteq \realset{}$. We must note that our developments can be extended to the multivariate case, i.e., $\mathcal{T} \subseteq \realset{D}$ with $D \geq 2$.}

We say that $Y$ is GP-distributed if any finite subset of random variables extracted from $Y$ has a joint Gaussian distribution \citep{Rasmussen2005GP}. By focusing on centered GP priors,\footnote{A GP $Z$ with mean function $\mu$ and kernel $k$ can be written in terms of a centered GP $Y$ with same kernel: $Z(\BfF) = \mu(\BfF) + Y(\BfF)$.} then $Y$ is completely defined by
\begin{equation}
	Y \sim \GP{0}{k},
	\label{eq:GP}
\end{equation}
where the kernel $k(\BfF,\BfF')$ evaluates the correlation between $Y(\BfF)$ and $Y(\BfF')$. For instance, $k(\BfF,\BfF') = 0$ if $Y(\BfF)$ and $Y(\BfF')$ are uncorrelated and is nonzero otherwise.

One of the main benefits of GPs relies on the tractability of conditional distributions. Let us consider $Y$ conditioned on an observation vector $\By_N = [y_1, \ldots, y_N]^\top$ evaluated at $(\BfF_1, \ldots, \BfF_N)$, where $\BfF_i = (f_{i,1}, \ldots, f_{i,Q})$ for $i = 1, \ldots, N$. Then, the conditional distribution $Y|\{\BY_N = \By_N\}$ for the Gaussian vector $\BY_N = [Y(\BfF_1), \ldots, Y(\BfF_N)]^\top$ is also Gaussian with conditional mean and conditional covariance functions given by
\begin{align}
	\mu(\BfF_\ast) &= \Bk^{\top}(\BfF_\ast) \BK^{-1} \By_N, 
	\label{eq:condGPEqs} \\
	c(\BfF_\ast, \BfF'_\ast) &= k(\BfF_\ast, \BfF'_\ast) - \Bk^\top(\BfF_\ast) \BK^{-1} \Bk(\BfF'_\ast),		
	\nonumber
\end{align}
with covariance matrix $\BK = (k(\BfF_i, \BfF_j))_{1 \leq i,j \leq N}$ and cross-covariance vector $\Bk(\BfF) = [k(\BfF, \BfF_1), \ldots, k(\BfF, \BfF_N)]^\top$. The conditional mean $\widehat{Y}(\BfF_\ast) = \mu(\BfF_\ast)$ is usually used as a point estimate of $Y(\BfF_\ast)$, and the conditional variance $v(\BfF_\ast) = c(\BfF_\ast, \BfF_\ast)$ is the expected square error of this estimate.

\subsubsection{Construction of stationary kernels for functional inputs}
\label{sec:functionalGPs:subsec:fkernels}

To establish proper kernels, we need to define the distance between functions. We consider the $L^2$-norm as it leads to simpler and closed-form expressions in Section \ref{sec:functionalGPs:subsec:projFun}:
\begin{equation}
	\|\BfR\|_{\Bell}^2 
	= \|\BfF - \BfF'\|_{\Bell}^2 
	=  \sum_{i = 1}^{Q} {\frac{\int_{\mathcal{T}} (f_i (t) - f'_i (t))^2 dt}{\ell_i^2}},
	\label{eq:fdist}
\end{equation}
where $\ell_{i} \in \realset{+}$ and $\int_{\mathcal{T}} (f_i (t) - f'_i (t))^2 dt < \infty$, for $i = 1, \ldots, Q$. The length-scale parameter $\ell_i$ can be viewed as a scale parameter for the $i$th functional input.

\begin{figure}[t!]
	\centering
	\begin{minipage}{0.53\columnwidth}
		\subfigure[\label{subfig:SE}SE]{\includegraphics[width=0.49\columnwidth]{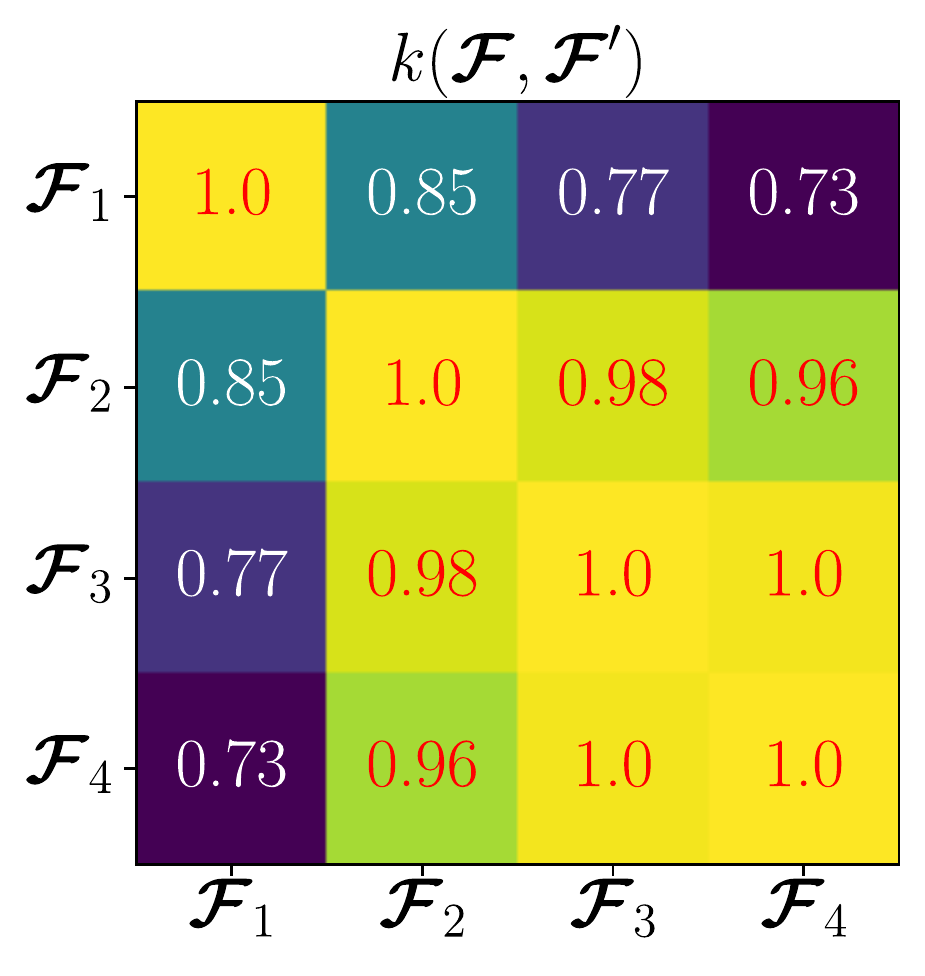}}
		\subfigure[\label{subfig:Mat52}Mat\'ern 5/2]{\includegraphics[width=0.49\columnwidth]{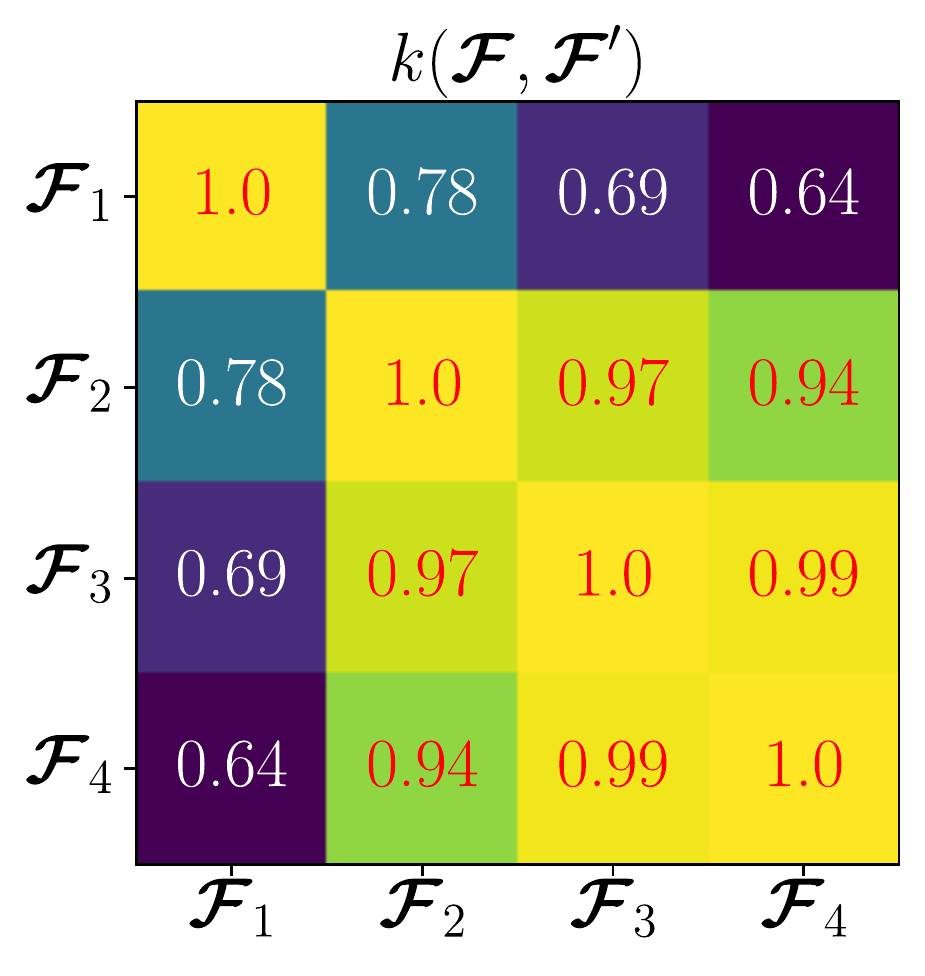}}
		\vskip-2ex
		
		\subfigure[\label{subfig:Mat32}Mat\'ern 3/2]{\includegraphics[width=0.49\columnwidth]{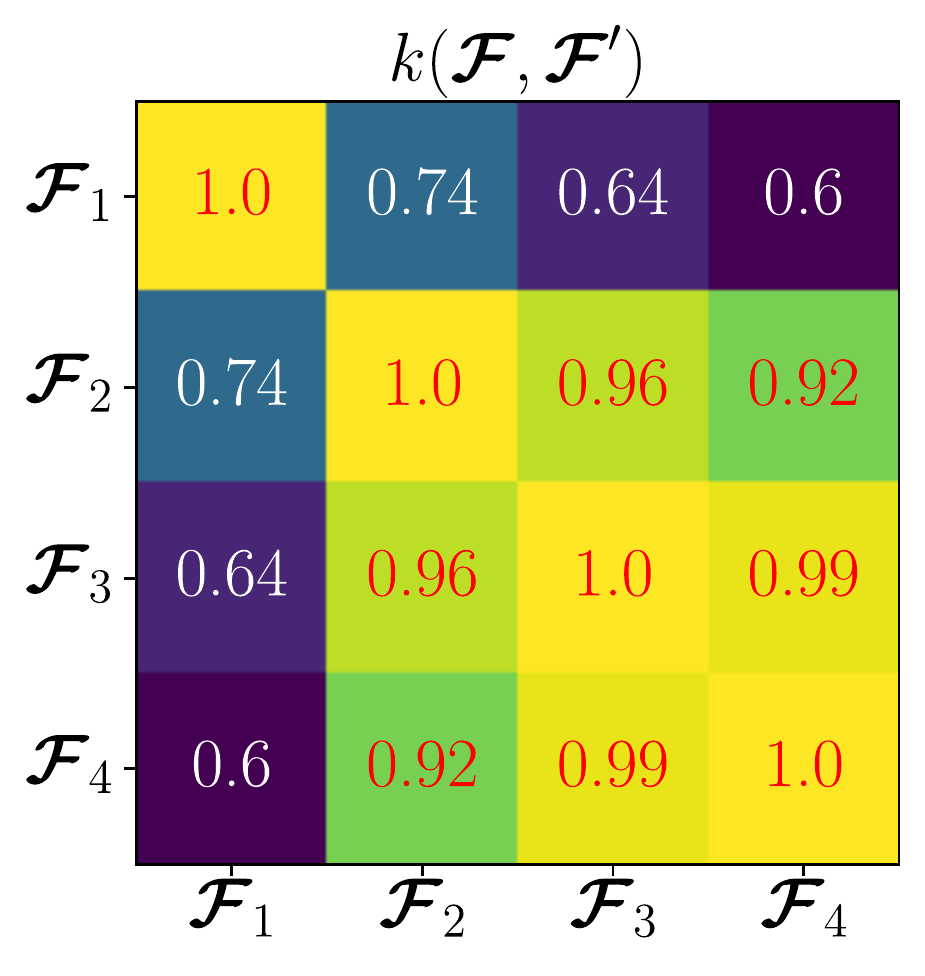}}
		\subfigure[\label{subfig:Exp}Exponential]{\includegraphics[width=0.49\columnwidth]{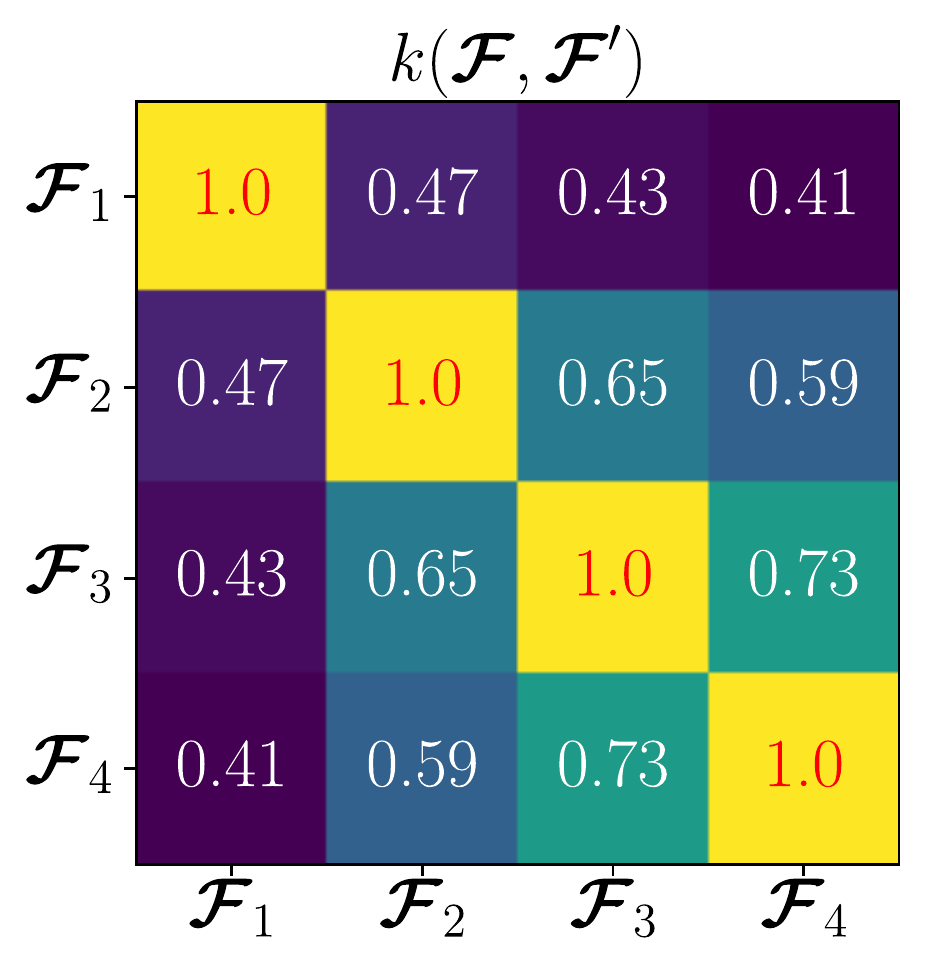}}
	\end{minipage}
	\begin{minipage}{0.45\columnwidth}
		\vskip2ex
		\centering
		\subfigure[\label{subfig:targetfuns}Target functions]{\includegraphics[width=\columnwidth]{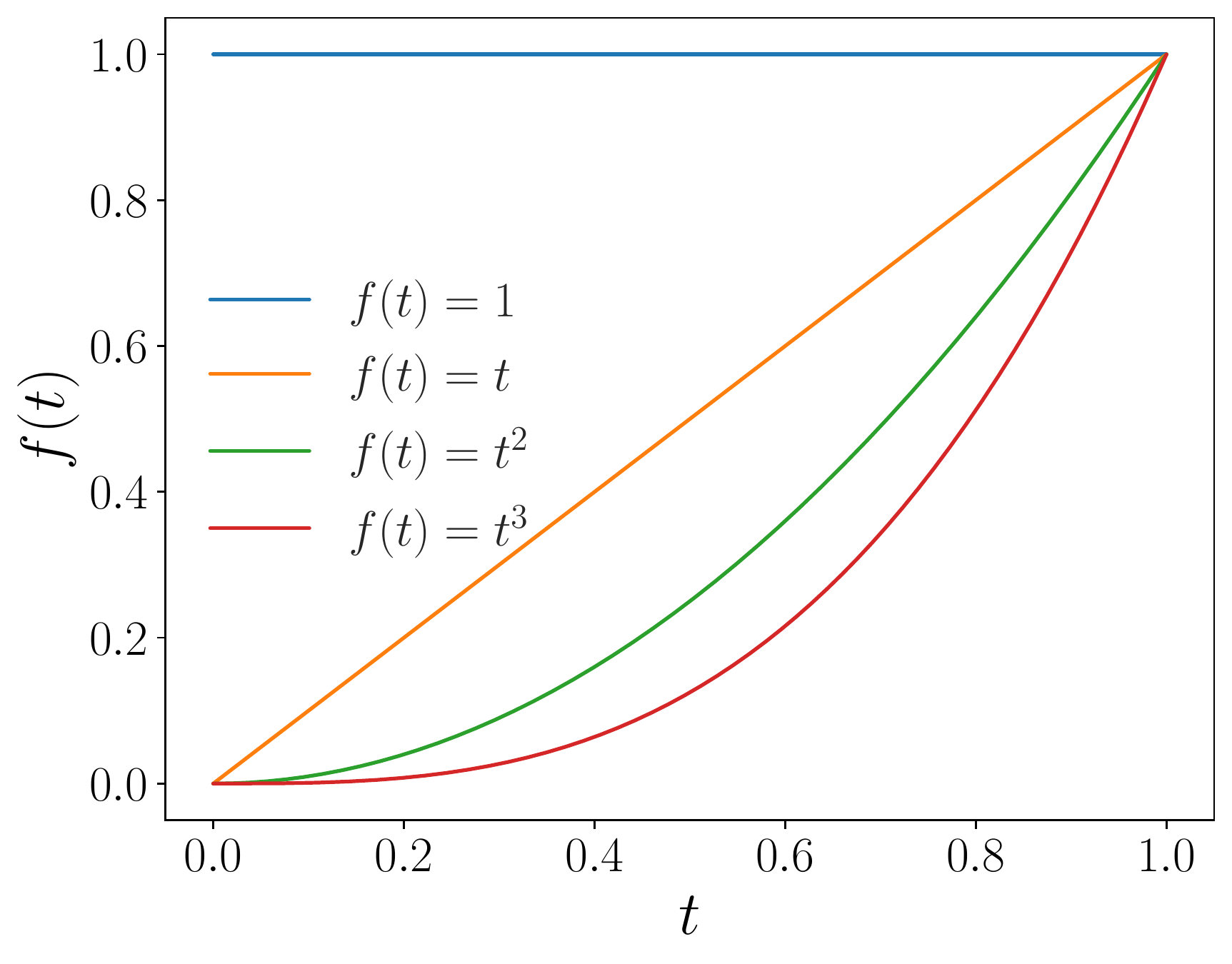}}
	\end{minipage}
	\caption{Effect of the kernels in Table \ref{tab:fkernels} with $\BfF_1 = (f(t) = 1)$, $\BfF_2 = (f(t) = t)$, $\BfF_3 = (f(t) = t^2)$ and $\BfF_4 = (f(t) = t^3)$ as functional inputs. The covariance parameters are set to $(\sigma^2 = 1, \ell = 1)$. The panels show the kernels (left) and the functions $f$ (right).
	}
	\label{fig:fkernels}
\end{figure}

Examples of valid stationary kernels are shown in Table \ref{tab:fkernels}, and illustrations of the effect of those kernels are displayed in Figure \ref{fig:fkernels}. We consider $\BfF_1 = (f(t) = 1)$, $\BfF_2 = (f(t) = t)$, $\BfF_3 = (f(t) = t^2)$ and $\BfF_4 = (f(t) = t^3)$ as inputs. We fix the variance parameter $\sigma^2 = 1$ and the length-scale parameter $\ell = 1$. {We can observe that the cross-covariance values $k(\BfF_i, \BfF_j)$, for $i,j = 2, 3, 4$ with $i \ne j$, are greater than those involving $k(\BfF_1, \BfF_j)$.} This results from the similarity between the functions $f(t)=t$, $f(t)=t^2$ and $f(t)=t^3$ on $\mathcal{T} = [0,1]$. Furthermore, from Figure \ref{subfig:SE} to \ref{subfig:Exp}, we note that GP models can be more or less sensitive to dissimilarities depending on the choice of the kernel. {For example, while the squared exponential (SE) kernel provides a correlation between $\BfF_2$ and $\BfF_4$ equal to one (Figure \ref{subfig:SE}), the exponential kernel suggests that the similarity between those inputs is weaker and equal to 0.73 (Figure \ref{subfig:Exp}).}

\begin{table}
	\centering
	\begin{tabular}{cc}
		\toprule
		Kernel & Mathematical expression $k_{\Btheta = (\sigma^2, \Bell)}(\BfR)$ \\
		\midrule
		Squared exponential (SE) & $\sigma^2 \exp\left\{ -\frac{1}{2} \|\BfR\|_{\Bell}^2 \right\}$ \\
		Mat\'ern 5/2 & $\sigma^2 \left( 1 + \sqrt{5} \|\BfR\|_{\Bell} + \frac{5}{3} \|\BfR\|_{\Bell}^2 \right) \exp\left\{ -\sqrt{5} \|\BfR\|_{\Bell} \right\}$ \\
		Mat\'ern 3/2 & $\sigma^2 \left( 1 + \sqrt{3} \|\BfR\|_{\Bell} \right) \exp\left\{ -\sqrt{3} \|\BfR\|_{\Bell} \right\}$ \\
		Exponential & $\sigma^2 \exp\left\{ -\|\BfR\|_{\Bell} \right\}$ \\				
		\bottomrule
	\end{tabular}
	\caption{Examples of valid stationary kernel functions }
	\label{tab:fkernels}
\end{table}

The exact computation of \eqref{eq:fdist} relies on its $Q$ integrals. Depending on the complexity of $f_i$ and $f'_i$, such integrals can be intractable. Furthermore, in many situations, only a finite number of evaluations of $f_i$ and $f'_i$ are available, but their functional structures are actually unknown {(e.g., for the case of discrete time series)}. For these reasons, functions are usually replaced by linear approximations where \eqref{eq:fdist} has a closed-form solution since the integrals operate over well-defined basis functions.

\subsubsection{Projection of functional inputs onto basis functions}
\label{sec:functionalGPs:subsec:projFun}

Consider the projection of $f$ onto a set of basis functions \citep{Ramsay2005functional}:
\begin{equation}
	f(t) \approx \sum_{j = 1}^{p} \phi_{j}(t) \alpha_{j} :=  g(t).
	\label{eq:basisConstr}
\end{equation}
Similarly, let $g'$ be the linear approximation of $f'$ with the same set of basis functions $\phi_{1}, \ldots, \phi_{p}$ and with the coefficients $\alpha'_{1}, \ldots, \alpha'_{p}$. Using \eqref{eq:basisConstr}, the integrals in \eqref{eq:fdist} are then approximated by the form $\int_{\mathcal{T}} (g_i(t) - g'_i(t))^2 dt$. {Note that in \eqref{eq:basisConstr}, for ease of notation, we intentionally dropped the subindex $i$ in the definition of $f_i$.}

Matricially, the integral in \eqref{eq:basisConstr} is given by
\begin{equation}
	\int_{\mathcal{T}} (g (t) - g' (t))^2 dt
	= \Bbeta^\top \BPsi \Bbeta,
	\label{eq:distFbasisMat}
\end{equation}
where $\BPsi = \int_{\mathcal{T}} \Bphi(t) \Bphi^\top(t) dt$, $\Bphi(t) = [\phi_{1}(t), \ldots, \phi_{p}(t)]^\top$, $\Bbeta = [\beta_{1}, \ldots, \beta_{p}]^\top$ and $\beta_{j} = \alpha_{j} - \alpha'_{j}$. Note from \eqref{eq:distFbasisMat} that the integral now operates over $\phi_{1}, \ldots, \phi_{p}$ whose functional structures are known.

For a wide range of basis families, such as in splines and PCA, the Gram matrix $\BPsi$ has an analytical form \citep[see, e.g.,][]{Ramsay2005functional,Shi2011functionalGPs}. In terms of computational savings, since $\BPsi$ does not depend on $f$ or $f'$, it can be computed only once, stored and reused when building up the kernels in Table \ref{tab:fkernels}. The consideration of orthogonal or orthonormal families of basis functions (e.g., in PCA) results in significant simplifications. For the former family, $\BPsi$ is a diagonal matrix given by $\BPsi = \operatorname{diag}(\int_{\mathcal{T}} \Bphi(t) \Bphi^\top(t) dt)$. For the orthonormal case, $\BPsi$ is the identity.

In this paper, we focus on projections based on PCA approximations as they lead to two main computational benefits \citep[see, e.g.,][for other implementations based on splines]{Betancourt2020fGPs}. First, $\BPsi = \BI$ due to the orthonormality of the basis functions. Second, the most relevant information from functional inputs is encoded in a few principal components. As an example, to satisfy a total inertia of $99.9\%$ for the 37 observation long discrete time series in Figure \ref{fig:toyExample8BRGMInputs}, PCA led to $\bm{p} = [1,4,3,3,2,6,4,9]$ where $p_i$ is the number of principal components for the $i$th functional input. Note that smaller values of $p_i$ are assigned to less varied functional profiles. We refer to Appendix \ref{app:projPCA} for a further discussion on the construction of the PCA basis functions.

{In the coastal flooding application, we consider discrete representations of the functional inputs. Therefore, we may be interested in exploiting classic measure distances such as the Euclidean or Mahalanobis distances; moreover, we may contemplate dedicated distance measures proposed for time series \citep[see, e.g.,][]{Gorecki2019DistanceTimeSeries,Mori2016DistanceMF}. Note that those distances will consider 37-dimensional input spaces compared to the $p$-dimensional spaces, where $p \in \{1, 9\}$, obtained by the PCA projection. The increase in the computational costs here can be mild, but the PCA approximation will be better in other situations where the input space generated by the discrete time series is large. For example, if time series contain hundreds or thousands of discrete values, it is preferable to use the PCA projection to achieve cheaper and faster computations.}

\subsection{Extension to spatial Gaussian processes}
\label{sec:spatGPs}
We now consider that $\{Y(\BfF, \Bx);  \BfF \in \mathcal{F}(\mathcal{T},\realset{})^Q, \Bx \in \realset{2}\}$ is a centered spatial GP with functional inputs $\BfF$ and spatial coordinates $\Bx = (x_1, x_2)$. Hence, GP $Y$ can be fully defined by constructing a valid kernel $k$ that accounts for both spatial information and functional inputs, i.e., $k((\BfF,\Bx), (\BfF',\Bx'))$. Note that if the approximations $g_1, \ldots, g_Q$ are considered instead, then $k((\BfF,\Bx), (\BfF',\Bx'))$ must be rewritten for $k((\BfG,\Bx), (\BfG',\Bx'))$. For the sake of consistency with Sections \ref{sec:functionalGPs:subsec:GPs} and \ref{sec:functionalGPs:subsec:fkernels}, we continue the discussion with the notations of $\BfF$ and $\BfF'$.

\subsubsection{Construction of the covariance function via separable kernels}
\label{sec:spatGPs:subsec:sepkernel}
A natural extension of GPs accounting for mixed variables relies on separable kernels defined by the product of subkernels \citep[see, e.g.,][]{Fricker2013GPNonseparableKernels,Roustant2020GroupKernels}. In our case, the kernel can be written as
\begin{equation}
	k((\BfF,\Bx), (\BfF',\Bx')) = k_f(\BfF, \BfF') k_x(\Bx, \Bx'),
	\label{eq:ksfproduct}
\end{equation}
where subkernels $k_x : \realset{2} \times \realset{2} \to \realset{}$ and $k_f : \mathcal{F}(\mathcal{T},\realset{})^Q \times \mathcal{F}(\mathcal{T},\realset{})^Q \to \realset{}$.\footnote{To avoid nonidentifiability of the variance parameters, $k_f$ is considered a correlation function, i.e., the variance is $\sigma_f^2 = 1$.} Note now that the kernel in 	\eqref{eq:ksfproduct} is attenuated by the spatial correlation. Therefore, although having similar functional inputs $\BfF$ and $\BfF'$, distant values of $\Bx$ and $\Bx'$ can result in small correlations. Since the process $Y$ remains a GP, the conditional formulas in \eqref{eq:condGPEqs} hold for \eqref{eq:ksfproduct} with tuples $(\BfF_i,\Bx_i)_{1 \leq i \leq N}$.

One of the main benefits of using \eqref{eq:ksfproduct} relies on the exploitation of Kronecker structures. In that case, we should consider tuples $(\BfF_i,\Bx_j)_{1 \leq i \leq R, 1 \leq j \leq S}$, where $R$ is the number of functional replicates (i.e. number of flood scenarios) and $S$ is the number of spatial points per map. This leads to a total of $N = R \times S$ observations. Denote the covariance matrices as $\BK_{f} = (k_f(\BfF_i, \BfF_j))_{1 \leq i,j \leq R}$ and $\BK_{x} = (k_x(\Bx_i, \Bx_j))_{1 \leq i,j \leq S}$. Then, from \eqref{eq:ksfproduct}, we have the Kronecker product $\BK = \BK_f \otimes \BK_x$; and the Cholesky factorization of $\BK$ given by $\BL = \BL_f \otimes \BL_x$, where $\BL_f$ and $\BL_x$ are the (lower triangular) Cholesky matrices of $\BK_f$ and $\BK_x$, respectively. This results in less expensive procedures since both Cholesky and inverse operations are applied on smaller matrices, reducing the computational complexity to $\mathcal{O}(R^3) + \mathcal{O}(S^3)$ \citep[compared to $\mathcal{O}(R^3 S^3)$ for standard implementations,][]{Alvarez2012kernelReview}. For large datasets such as the one detailed in Section \ref{sec:BRGMapp}, computing either $\BK$ or $\BL$ (or their inverses) can easily run out of memory as either $R$ or $S$ increases. To mitigate this drawback, more efficient computations can be obtained by solving triangular-structured linear systems rather than directly computing those matrices \cite[see, e.g.,][]{Bilionis2013MoGPs}. For easy self-reading, in Appendix \ref{app:kronGPs}, we summarize the main Kronecker-based computations used in this paper.

\subsubsection{Connection to other GP developments}
\label{sec:spatGPs:subsec:otherGPdevs}

\paragraph{Linear models of coregionalization (LMC).}
The process $Y$ can be written as a multioutput process $Z$ where the outputs are driven by a given set of functions, i.e., $Z_{i}(\Bx) := Y(\BfF_i, \Bx)$, for $i = 1, \cdots, R$. In that case, \eqref{eq:ksfproduct} yields:
\begin{align}
	k_{i,j}(\Bx, \Bx') = b_{i,j} \; k_x(\Bx, \Bx'),
	\label{eq:ksfmo}
\end{align}
where $b_{i,j} := k_f(\BfF_i, \BfF_j)$ for $i,j = 1, \ldots, R$. The kernel in \eqref{eq:ksfmo} follows a similar structure to the one in LMC, more precisely, to the one in intrinsic coregionalization models \citep[ICMs,][]{Alvarez2012kernelReview}. Note that $k_f$ involves only the estimation of $Q$ length-scale parameters rather than the $R(R+1)/2$ coefficients ($R \gg Q$) corresponding to the upper triangular block of the symmetric coregionalization matrix $\BB = (b_{i,j})_{1 \leq i,j \leq R}$. Another benefit of considering $\BK_{f}$ is its positive definiteness condition since $k_f$ is defined as a kernel. Since this condition is not necessarily satisfied by $\BB$, ICMs may lead to numerical instabilities due to noninvertible matrices. In practice, $\widetilde{\BB} = \BB + \operatorname{diag}([\kappa_1, \ldots, \kappa_R])$ is considered to ensure the positive definiteness condition, but this implies the estimation of the $R$ parameters $\kappa_1, \ldots, \kappa_R$.

\paragraph{Sparse approximations.}
Further simplifications are obtained by sparse approximations \citep{Titsias2009SparseGPs,Hensman2013GPbigData}. Recall that $Z_i(\Bx) = Y(\BfF_i, \Bx)$ for $i = 1, \ldots, R$. The distribution of $(Z_1, \ldots, Z_R)$, conditioned to $(Z_i(\Bx_{1}) = z_{i,1}, \ldots, Z_i(\Bx_{S}) = z_{i,S})_{1 \leq i \leq R}$, can be approximated by a cheaper but tractable variational distribution conditioned to $(Z_i(\Bu_1) = \widetilde{z}_{i,1}, \ldots, Z_i(\Bu_M) = \widetilde{z}_{i,M})_{1 \leq i \leq R}$, with $M \ll S$. Then, using a low rank approximation of $k_s$, the complexity $\mathcal{O}(R^3) + \mathcal{O}(S^3)$ is reduced to $\mathcal{O}(R^3) + \mathcal{O}(S M^2)$. The inducing variables $\Bu_1, \ldots, \Bu_M$ are estimated via variational inference \citep[see, e.g.,][]{Hensman2013GPbigData}, and $M$ is fixed seeking a trade-off between the computational complexity and approximation quality. While a large value of $M$ leads to more precise but expensive models, a small value results in faster but less accurate approximations.

\paragraph{Variational inference (VI).}
As our functional framework preserves the structure of ICMs, it can be easily plugged into other types of multioutput GPs (MoGPs) based on VI \citep[see, e.g.,][]{Hensman2013GPbigData,Moreno2018HeteroMoGPs,GPflow2020multioutput}. As an example, our functional framework can be fitted via stochastic VI (SVI), which scales well for large values of $S$ \citep{Hensman2013GPbigData}. Considering a separable Kronecker-based kernel and applying SVI (under sparse assumptions) only to the spatial kernel, the complexity of the resulting GP decreases to $\mathcal{O}(R^3) + \mathcal{O}(B M^2)$ with batch size $B \ll S$. The value of $B$ is manually fixed depending on the available storage capacity. More efficient variational implementations are obtained by using interdomain approximations \citep[see][]{Lazaro2009InterdomainGPs,GPflow2020multioutput}. As another example, heterogeneous MoGPs can also be established by considering that each output has its own likelihood function \citep[e.g. a Gaussian, a Bernoulli or a Poisson likelihood,][]{Moreno2018HeteroMoGPs}.

Since our MoGP implementations are based on the \texttt{GPflow} Python library, they enjoy a great variety of dedicated VI developments \citep[see the documentation in][]{GPflow2017}, including those for the SVI and heterogeneous MoGPs. Although the variational features in \citep{Moreno2018HeteroMoGPs} are not exploited in this paper, they are available for the scientific community for further research (see Section \ref{sec:numillustration:subsec:codes} for further details).

\section{Results and Discussions}
\label{sec:results}

\subsection{Python and R implementations}
\label{sec:numillustration:subsec:codes}
The \texttt{GPflow} Python library relies on variational inference to meet the twin challenges of nonconjugacy and scale \citep{GPflow2017}. To the best of our knowledge, \texttt{GPflow} does not support Kronecker-based composite kernels in its latest release. Therefore, implementing Kronecker-structured objects requires significant modifications at the root level. This motivates developments based on the \texttt{kergp} R package \citep{Deville2019kergp}. The \texttt{kergp} package is not equipped with sparse-variational approximations as \texttt{GPflow} is. Nevertheless, the \texttt{kergp} package is sufficiently flexible to account for both functional data and Kronecker-structured composite kernels. Both Python and R implementations are available on GitHub: \url{https://github.com/anfelopera/spatfGPs}.

{
	
	\subsection{Performance indicators}
	\label{sec:numillustration:subsec:indicators}
	We consider the three following performance indicators for the predictions: the root mean square error (RMSE), the $\mathcal{Q}^2$ value and the coverage accuracy ($\mbox{CA}$). The first two assess the quality of the predictive mean, and the CA evaluates the quality of the predictive variance. The $\mathcal{Q}^2$ criterion is given by
	\begin{equation}
		\mathcal{Q}^2 = 1 - \frac{\sum_{i=1}^{N_{\text{test}}} (y_i - \widehat{y}_i)^2}{\sum_{i=1}^{N_{\text{test}}} (y_i - \overline{y})^2},
		\label{eq:Q2}
	\end{equation}
	where $\widehat{y}_1, \ldots, \widehat{y}_{N_{\text{test}}}$ and $\overline{y}$ are the predictions and the average of test data $y_1, \ldots, y_{N_{\text{test}}}$, respectively. For noise-free observations, $\mathcal{Q}^2$ is equal to one if predictions are exactly equal to the test data, zero if they are equal to $\overline{y}$, and negative if they perform worse than $\overline{y}$.
	
	The $\mbox{CA}$ of the $c$-standard deviation confidence intervals, denoted as $\mbox{CA}_{\pm c\sigma}$, indicates the proportion of the test data that is contained in the confidence intervals:
	\begin{equation}
		\mbox{CA}_{\pm c\sigma} = \sum_{i=1}^{N_{\text{test}}} \mathds{1}_{y_i \in [\widehat{y}_i - c \sigma, \widehat{y}_i + c \sigma]},
		\label{eq:CA}
	\end{equation}
	where $\sigma^2$ is the predictive variance provided by the GP model, and $\mathds{1}_{y_i \in [\widehat{y}_i - c \sigma, \widehat{y}_i + c \sigma]}$ is equal to one if $y_i \in [\widehat{y}_i - c \sigma, \widehat{y}_i + c \sigma]$ and zero otherwise.
}

\subsection{Numerical illustrations on synthetic examples}
\label{sec:numillustration}
We test the proposed GP framework under different situations depending on the data availability. First, we consider the case where different design points per map are considered. To simplify the multioutput learning task, strong correlations between spatial events and functional inputs are assumed. Second, highly variable maps are predicted considering Kronecker-structured design points. Finally, we apply the model for predicting unobserved events given a set of learning events (forecasting task).

\subsubsection{Multioutput illustration}
\label{sec:numillustration:subsec:2Dtoy}

{
	\paragraph{Synthetic dataset.}
	We consider a coupled system consisting of $R=20$ outputs and $Q=8$ inputs. To emulate functional patterns, the latter functions are sampled from the predefined GPs given by $f_{i} \sim \GP{\mu_{i}}{k_o}$ for $i = 1, \ldots, Q$, with mean function $\mu_{i}$ and Mat\'ern 5/2 kernel $k_o$. We fix the variance $\sigma_o^2 = 2.5\times10^{-3}$ and the length-scale $\ell_{o} = 0.8$ seeking small variations between $f_{i}$ and $\mu_{i}$. The mean functions $\mu_{1}, \ldots, \mu_{Q}$ are also GP realizations, i.e., $\mu_{i} \sim \GP{0}{k_i}$, with Mat\'ern 5/2 kernels, variances $\sigma_{\mu_i}^2 = \frac{1}{2}$ and length-scales $\ell_{\mu_i} = \frac{i}{10}$. Note that $f_{i}$ becomes less variable as $\ell_{\mu_i}$ increases (Figure \ref{fig:toy2DMaps}). Then, we sample $100 \times 100$ maps using a 2D Mat\'ern 5/2 kernel with variance $\sigma_x^2 = 1$ and length-scales $\ell_{x,1} = \ell_{x,2} = 0.2$. To correlate the inputs, a Mat\'ern 5/2 kernel is used with length-scales $\ell_{f,i} = 2$ for $i = 1, \cdots, 8$. For convenience, strong correlations are considered to determine sampled maps that resemble each other (Figure \ref{fig:toy2DMaps}).
	\begin{figure}[t!]
		\centering
		\includegraphics[width=\columnwidth]{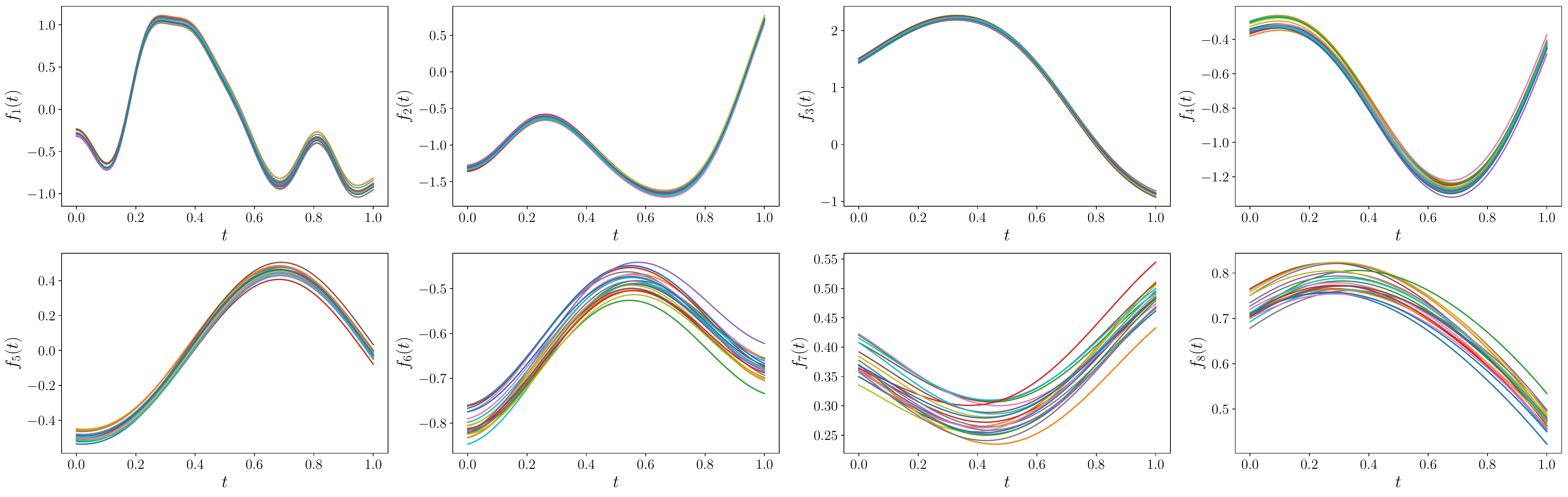}
		
		\medskip
		
		\foreach \i in {1,10,19} {
			\includegraphics[width=0.32\columnwidth,height=0.3\columnwidth]{toyExample2GPflowSVGPCoregionMap\i}
		}
		\caption{(top) Sampled replicates for the (8) inputs and (bottom) examples of sampled spatial events (20) used in the illustration of Section \ref{sec:numillustration:subsec:2Dtoy}.}
		\label{fig:toy2DMaps}
	\end{figure}
	
	\paragraph{Exact multioutput learning.}
	We now compare the performance of the proposed functional MoGP (denoted as fMoGP) to the performance of the standard MoGP via LCM. For the MoGP, we use the developments available in \texttt{GPflow} \citep{GPflow2017}. Both models consider Mat\'ern 5/2 kernels with the same initial spatial parameterization: $\sigma_{x,o}^2=1$ and $\ell_{x,1,o}=\ell_{x,2,o}=0.5$. For the MoGP, as suggested by \citet{GPflow2017}, we randomly initialize the coefficients of $\BB$. For the fMoGP, a Mat\'ern 5/2 kernel is used with initial length-scales $\ell_{f,i,o} = 0.5$ for $i = 1, \ldots, 8$. Then, the hyperparameters and covariance parameters of both models are estimated via maximum likelihood (ML) using $(\sigma_{x,o}^2, \ell_{f,1,o}, \ldots, \ell_{f,8,o})$ as the initial values of a gradient-based optimization. The gradients are computed using the automatic differentiation tool from \texttt{GPflow}.
	
	In the training step, we consider different maximin Latin hypercube designs (LHDs) with $S=35$ spatial points per map, leading to a total of 700 training points.\footnote{The budget has been fixed considering the computational capacity of an Intel(R) Core(TM) i5-4300M CPU @ 2.60GHz and 8 Gb RAM.} We use the enhanced stochastic evolutionary (ESE)-based LHD implementation in the \texttt{SMT} Python library \citep{SMT2019,Jin2005LHS}. Note that since tensorized designs are not considered, Kronecker-structured models cannot be exploited.
}

\begin{figure}[t!]
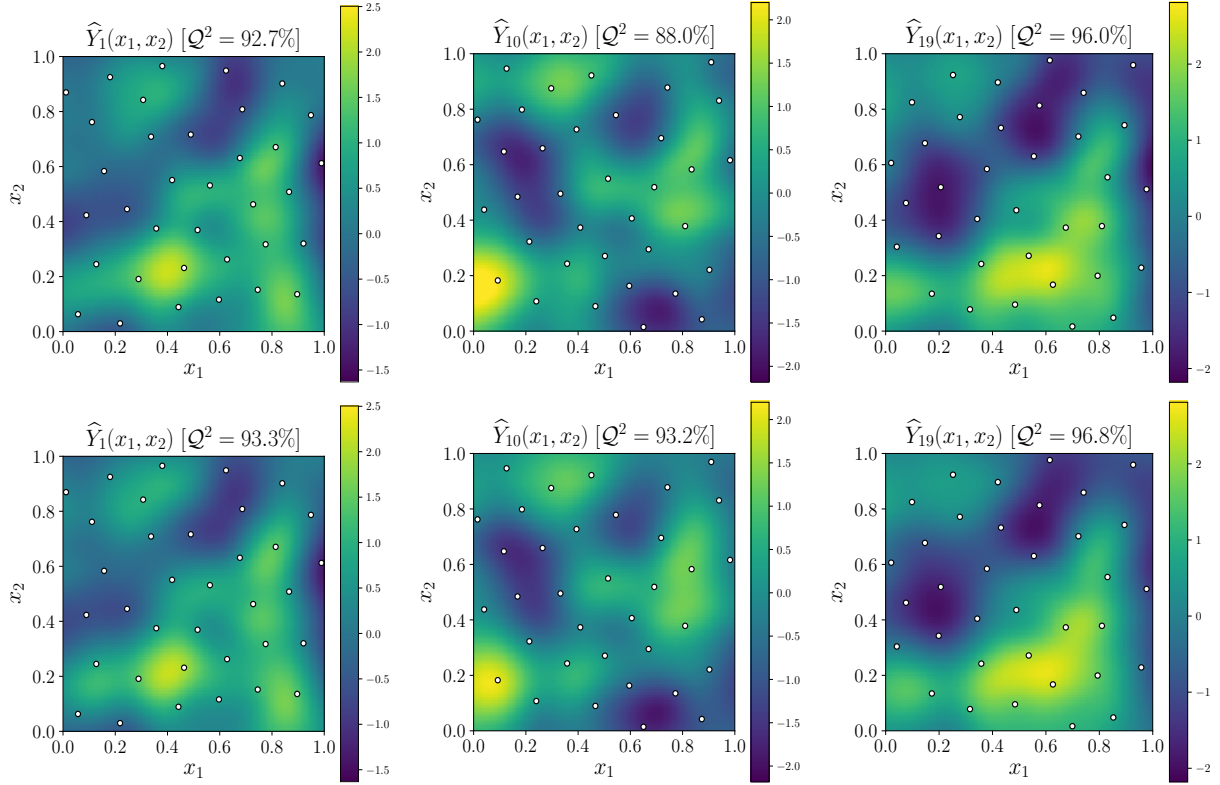

	\foreach \i in {1,10,19} {
		\begin{minipage}{0.32\columnwidth}	
			\centering				
			\includegraphics[width=\columnwidth,height=\columnwidth]{toyExample2GPflowSVGPCoregionMapPred\i}
			
			\includegraphics[width=\columnwidth,height=\columnwidth]{toyExample2GPflowSVGPCoregionMapPredf\i}
		\end{minipage}
	}
	\caption{Predictions for the spatial events in Figure \ref{fig:toy2DMaps}. The results are shown using either the standard MoGP via LMC (top) or the proposed functional MoGP framework (bottom). The white dots represent the spatial design points. The resulting $\mathcal{Q}^2$ value per map is also shown on top of each panel.}
	\label{fig:toy2DPredMaps}
\end{figure}

{
	\paragraph{Results.}
	Figure \ref{fig:toy2DPredMaps} shows the predictions provided by both MoGP and fMoGP. We can observe that both models capture the spatial dynamics of the maps in Figure \ref{fig:toy2DMaps}, leading to $\mathcal{Q}^2$ values of approximately $0.943$ and $0.960$ (averaged values over the 20 maps), respectively. Here, the $\mathcal{Q}^2$ criterion is computed over an equispaced $100\times100$ grid fixed for all the maps $Y_1, \ldots, Y_{20}$. After repeating this experiment using ten different LHDs, the MoGP and fMoGP led to $\mathcal{Q}^2 = 93.2 \pm 1.4$ \% (mean $\pm$ standard deviation) and $\mathcal{Q}^2 = 95.9 \pm 0.6$ \%, respectively. This raises the conclusion that accounting for the functional structure in the coregionalization matrix results in prediction improvements.}

\begin{figure}[t!]
	\centering
	\includegraphics[width=0.37\columnwidth]{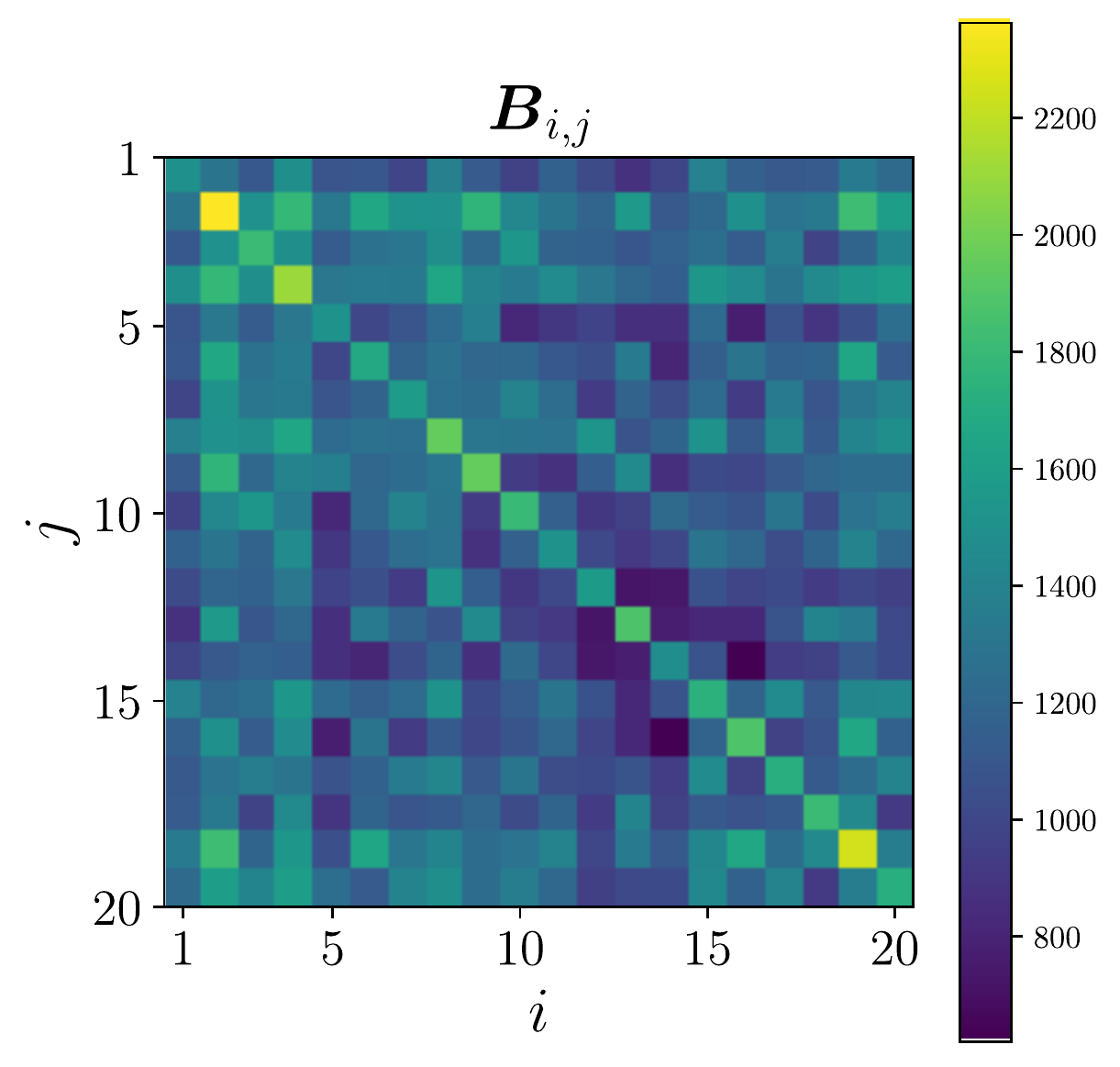}
	\hskip5ex
	\includegraphics[width=0.37\columnwidth]{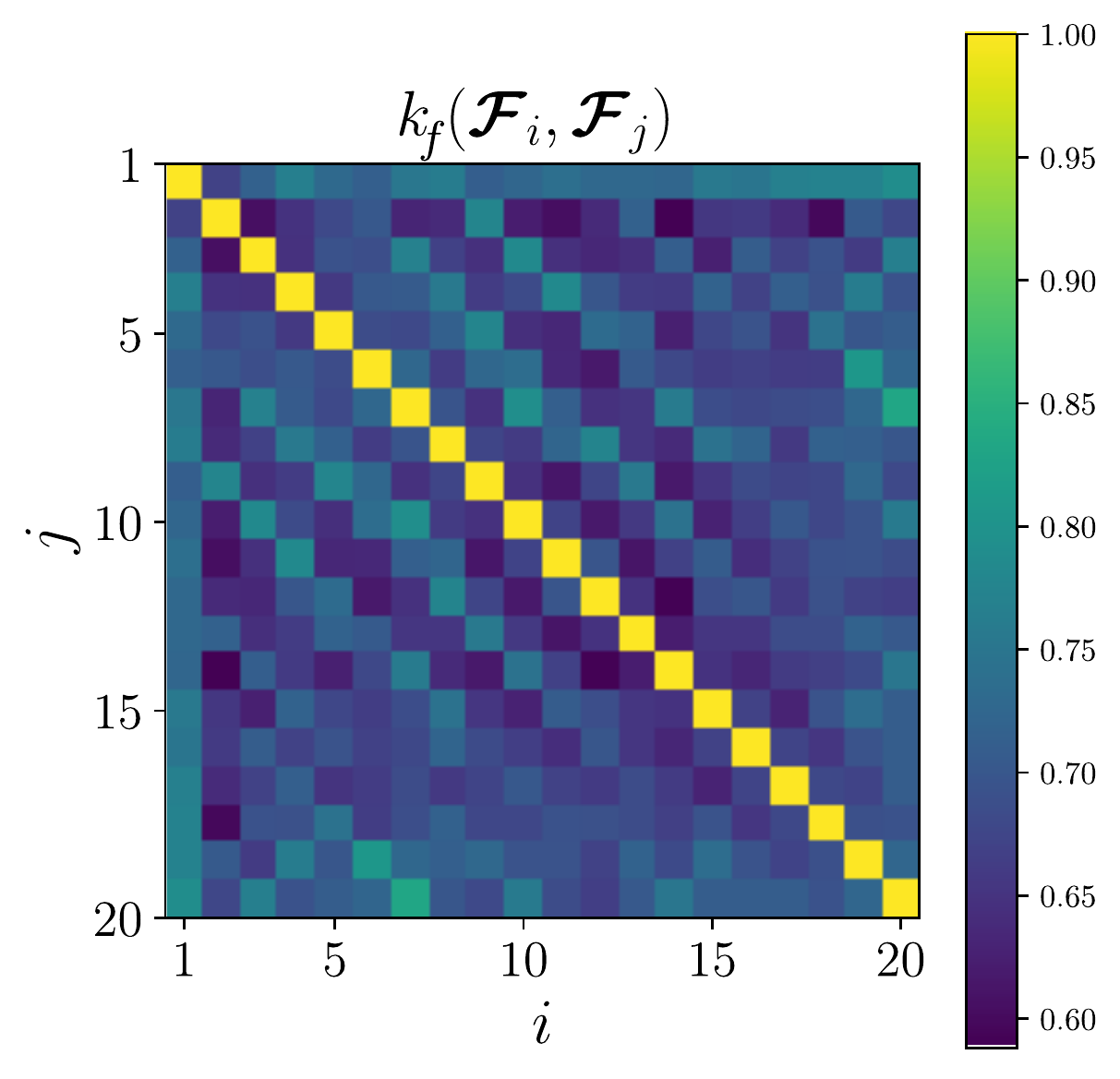}
	\caption{Coregionalization matrices for the MoGP (left) and fMoGP (right) models used in Section \ref{sec:numillustration:subsec:2Dtoy}.}
	\label{fig:toy2DCoregionKernel} \vskip-1.5ex
\end{figure}

Although both matrices $\BB$ and $\BK_f = (k_f(\BfF_i, \BfF_j))_{1 \leq i,j \leq R}$ exhibit the strongest correlations in the diagonal (Figure \ref{fig:toy2DCoregionKernel}), $\BB$ led to extremely high values. This affects the estimation of the variance, leading to $\widehat{\sigma}_{x,\text{MoGP}}^2=5.8\times 10^{-4}$ compared to $\sigma_x^2=1$ (true variance) and $\widehat{\sigma}_{x,\text{fMoGP}}^2=0.96$. For the length-scales of $k_x$, models led to $\hat{\Bell}_{x,\text{MoGP}} = (0.20,0.22)$ and $\hat{\Bell}_{x,\text{fMoGP}} = (0.20, 0.21)$ compared to the true results fixed to $\Bell_x = (0.2,0.2)$. For the length-scales of $k_f$, the fMoGP estimated $\hat{\Bell}_f = (2.57,1.84,1.83,1.66,1.74,3.55,1.46,1.53)$ compared to the true results fixed to 2.

\subsubsection{Sparse-variational illustration}
\label{sec:numillustration:subsec:2DtoySVI}

{
	\paragraph{Synthetic dataset.}
	The synthetic dataset is generated by following the sampling scheme used in Section \ref{sec:numillustration:subsec:2Dtoy} but with some covariance parameters slightly changed to find more variable spatial events. We fix $\sigma_{o}^2 = 4\times10^{-2}$ and $\ell_{x,1,o} = \ell_{x,2,o} = 0.1$. We then sample 50 maps using a $100 \times 100$ grid. Examples of the generated spatial events are shown in Figure \ref{fig:toy2DMapsLCM}.
	\begin{figure}
		\centering
		\foreach \i in {1,30,50} {
			\includegraphics[width=0.32\columnwidth,height=0.3\columnwidth]{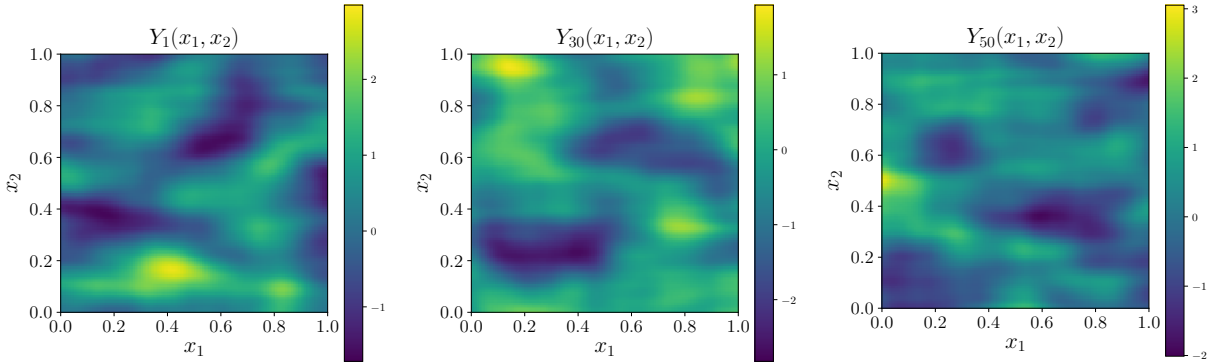}
		}
		\caption{Examples of the sampled maps used in the illustration of Section \ref{sec:numillustration:subsec:2DtoySVI}.}
		\label{fig:toy2DMapsLCM}
	\end{figure}
	
	\paragraph{Approximate multioutput learning.}
	We have adapted the sparse-variational MoGP (sv-MoGP) framework in \citep{GPflow2020multioutput} to account for functional inputs. We denote the resulting model as sv-fMoGP. The efficiency of sv-MoGP relies on considering sparse approximations together with dedicated SVI schemes and tensor-structured data.
	
	For the spatial design points, we use a maximin LHD of $S = 500$ points. This results in $N = 25 \times 10^3$ training data, a significantly larger amount of data compared to the nontensorized illustration proposed in Section \ref{sec:numillustration:subsec:2Dtoy}. The 50 spatial events are assumed to share the same inducing variables.\footnote{This assumption can be relaxed to address different sets of inducing variables per map but at the cost of complex inference.} Those variables are initialized as a maximin LHD seeking to cover the spatial space. To assess the influence of the inducing variables, we test models where the value of $M$ is gradually increased, i.e., $M = 20, 50, 100, 200$. In order to avoid running out of memory,\footnote{The experiments here have been executed on a single core of an Intel(R) Core(TM) i5-4300M CPU @ 2.60 GHz and 8 Gb RAM.} we apply the SVI with a minibatch size of $B = 200$. Both the covariance and variational parameters are estimated via SVI based on gradient-based optimization using automatic differentiation \citep[see][]{Hensman2013GPbigData,GPflow2020multioutput}.
	
	\paragraph{Results.}
	We tested both sv-MoGP and sv-fMoGP using ten different designs with $S = 500$ spatial points. The $\mathcal{Q}^2$ results (mean $\pm$ standard deviation) are shown in Table \ref{tab:toy2DPredPerformanceLCM}. The $\mathcal{Q}^2$ criterion was computed considering the spatial locations that were not used for training the models for all 50 maps. From Table \ref{tab:toy2DPredPerformanceLCM}, we observe that sv-fMoGP outperformed sv-MoGP, leading to absolute $\mathcal{Q}^2$ improvements greater than 12\%. We can also note that sv-fMoGP resulted in accurate predictions for $M \geq 100$.
	\begin{table}[t!]
		\centering
		\caption{Multioutput learning performance of sv-MoGP and sv-fMoGP. The $\mathcal{Q}^2$ results (mean $\pm$ standard deviation) are computed considering 10 replicates of the experiment in Section \ref{sec:numillustration:subsec:2DtoySVI}.}
		\label{tab:toy2DPredPerformanceLCM}	
		\begin{tabular}{ccccc}
			& \multicolumn{4}{c}{$\mathcal{Q}^2$ [\%] (mean $\pm$ standard deviation)} \\
			\toprule				
			\multirow{2}{*}{Model}	& \multicolumn{4}{c}{Number of Inducing Variables} \\
			& $M=20$ & $M=50$ & $M=100$ & $M=200$ \\
			\midrule
			sv-MoGP  & $53.6\pm 3.2$ & $75.1\pm 2.1$ & $81.4\pm 1.5$ & $84.0\pm 0.8$ \\
			sv-fMoGP & \boldsymbol{$69.9\pm 1.7$} & \boldsymbol{$87.4\pm 0.9$} & \boldsymbol{$93.6\pm 1.2$} & \boldsymbol{$96.6\pm 0.1$} \\
			\bottomrule	
		\end{tabular}
	\end{table}
	
	Although the predictability of both models improves as $M$ increases, fitting them to training data becomes costly. While for $M = 100$ the CPU times for $4 \times 10^3$ gradient evaluations of the SVI were approximately 0.32 h and 0.74 h (for sv-MoGP and sv-fMoGP, respectively), for $M = 200$, they were approximately 0.96 h and 1.85 h, respectively. sv-fMoGP was much slower since the gradients with respect to the length-scales $\ell_{f,1}, \ldots, \ell_{f,8}$ were more expensive to evaluate.
}

\subsubsection{Inference of unobserved outputs}
\label{sec:numillustration:subsec:2DtoyKronecker}

{
	\paragraph{Synthetic dataset.}
	We consider a coupled system consisting of $R=1001$ outputs and $Q=8$ functional inputs. We sample the latter functions as proposed in Section \ref{sec:numillustration:subsec:2Dtoy} but consider centered processes: $f_i \sim \GP{0}{k_i}$ with Mat\'ern 5/2 kernels, $\sigma_i^2 = \frac{1}{2}$ and $\ell_{i} = \frac{i}{10}$ for $i = 1, \ldots,q$. We generate the corresponding 1001 maps using an equispaced grid of $10 \times 10$ spatial locations. This leads to $S=100$ design points per map. We use the same kernel parameterization proposed in Sections \ref{sec:numillustration:subsec:2Dtoy} and \ref{sec:numillustration:subsec:2DtoySVI}.
	
	\paragraph{Inference of new maps.}
	As discussed in Section \ref{sec:BRGMapp}, our goal is to predict the consequence (i.e., the map of maximum water level $H_\max$) of unobserved storm events. This is achieved by correlating the hydrometeorological drivers (functional inputs). The length-scales $\ell_1, \ldots, \ell_{Q}$ of kernel $k_f$ can be estimated using data from the observed events. Therefore, since the structure of $k_f$ is completely defined after fitting the GP, our model can be applied for forecasting purposes. The framework in \citep{GPflow2020multioutput} requires the estimation of the mean and covariance functions of the variational distribution, which can only be estimated for observed outputs. This makes both sv-MoGP and sv-fMoGP inapplicable to forecasting tasks. However, we can still exploit Kronecker-based composite kernels using \texttt{kergp}.\footnote{Experiments here are executed on a single core of an HP cluster with an Intel biprocessor, 32$\times$2.2 GHz cores, 64 GB RAM.}
	
	In this example, we are interested in the inference of the 1001st map using data from the first 1000 maps. The performance of the model is assessed in terms of the number of learning maps $R$. Here, the covariance parameters are estimated via ML using the derivative-free constrained optimizer by linear approximations (COBYLA) implemented in the \texttt{NLOpt} package \citep{JohnsonNLopt}. We have also tested gradient-based optimizers, but they led to more expensive procedures.
	
	\paragraph{Results.}
	Figure \ref{fig:toyExample4NewMapConfInt} shows predictions using one standard deviation confidence intervals. We can observe that the predictive mean becomes closer to the test data as $R$ increases, with a $\mathcal{Q}^2$ improvement of 24.1\% between the results using 5 and 1000 learning events. Moreover, the model only needed $R = 200$ learning events to achieve accurate predictive performance of $\mathcal{Q}^2\sim 90\%$. In terms of the uncertainties, the predictive intervals cover the test data and decrease as $R$ increases.
	\begin{figure}
		\centering	
		\includegraphics[width=\columnwidth]{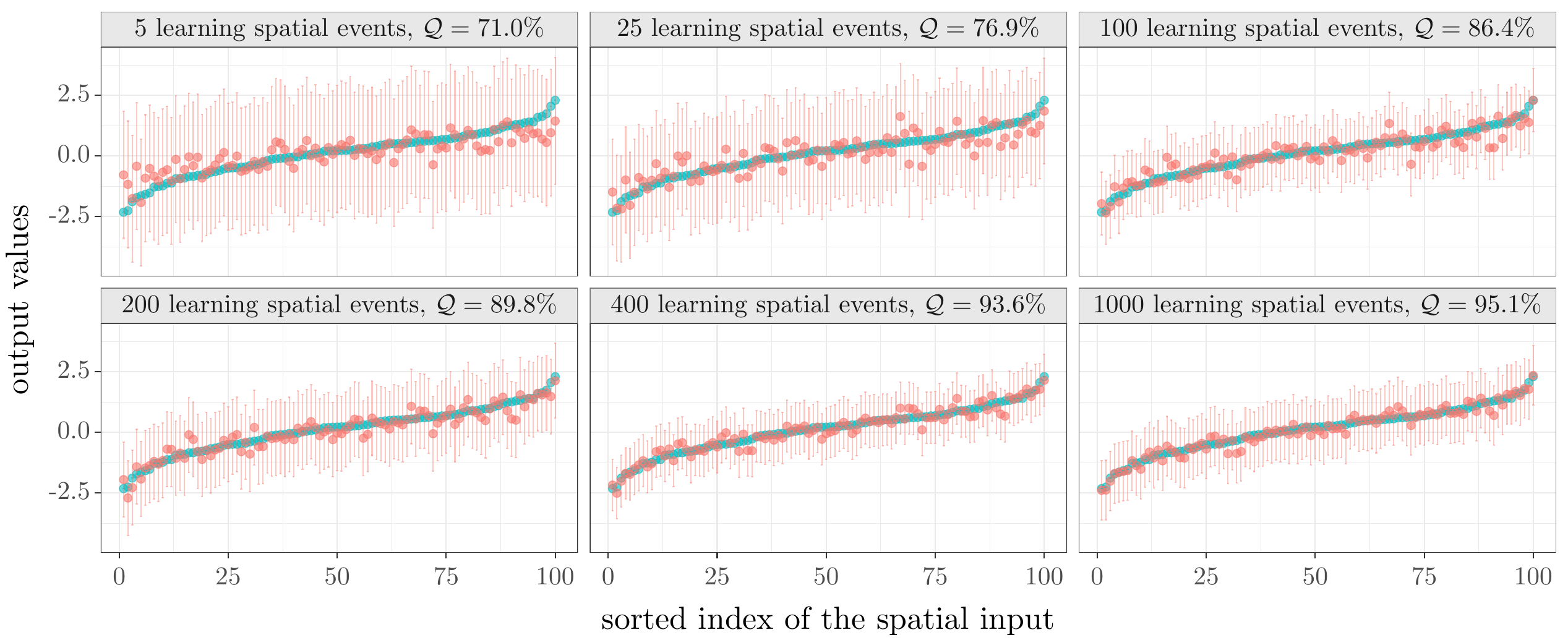}
		\caption{Predictions from Section \ref{sec:numillustration:subsec:2DtoyKronecker} considering one standard deviation confidence intervals. Each panel shows the following: the sampled test data (blue dots), the predictive mean (red dots) and the confidence intervals (red whiskers). Outputs are sorted by increasing order of the sampled test data.}
		\label{fig:toyExample4NewMapConfInt}
	\end{figure}
}


\subsection{Coastal flooding application}
\label{sec:BRGMresults}
We now apply our framework to the coastal flooding application described in Section \ref{sec:BRGMapp}. As we focus on forecasting purposes, we use our R implementations based on \texttt{kergp} \citep{Deville2019kergp}.

\subsubsection{Numerical settings}
\label{sec:BRGMresults:subsec:settings}

\paragraph{Data preprocessing.} Some of the hydrometeorological forcing conditions, such as the wave peak direction $\Dp$ and the wind direction $\Du$, are defined using nautical conventions (i.e., with north considered the reference). Then, as $\Du$ represents an angle, a slight change in winds coming from the north may lead to high angle variations and therefore to complex PCA representations. To avoid this drawback, in our experiments, we replace the tuples $(\Hs, \Dp)$ and $(\U, \Du)$ with the Cartesian tuples $(\Hs_x, \Hs_y)$ and $(\U_x, \U_y)$ given by $\Hs_x = \Hs \cdot \sin(\Dp)$, $\Hs_y = \Hs \cdot \cos(\Dp)$, $\U_x = \U \cdot \sin(\Du)$ and $\U_y = \U \cdot \cos(\Du)$. This results in a set of hydrometeorological functional inputs consisting of ($\msl$, $\tide$, $\surge$, $\Tp$, $\Hs_x$, $\Hs_y$, $\U_x$, $\U_y$).

\begin{figure}[t!]
	\centering
	\begin{minipage}{0.32\textwidth}
		\begin{tikzpicture}
			\node[anchor=south west,
			xshift=-\textwidth,
			yshift=-\textwidth] (image) at (current page.south west) {
				\includegraphics[width=\textwidth]{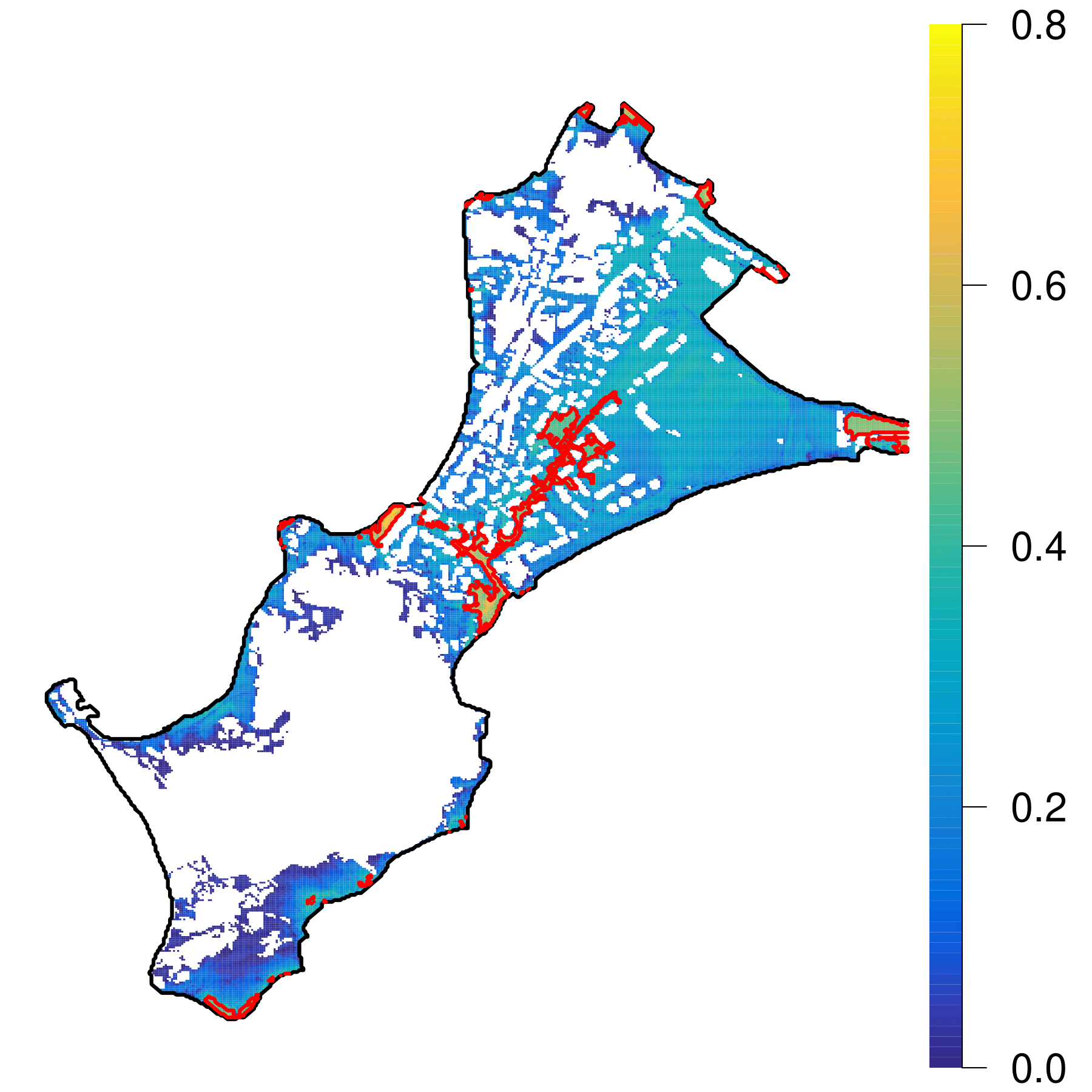}
			};
			\draw[->] (-\textwidth,-\textwidth)--++(0:1) node[right]{\footnotesize$x_1$};
			\draw[->] (-\textwidth,-\textwidth)--++(90:1) node[above]{\footnotesize$x_2$};
		\end{tikzpicture}
	\end{minipage}
	\hskip5ex
	\begin{minipage}{0.52\textwidth}
		\includegraphics[width=\textwidth]{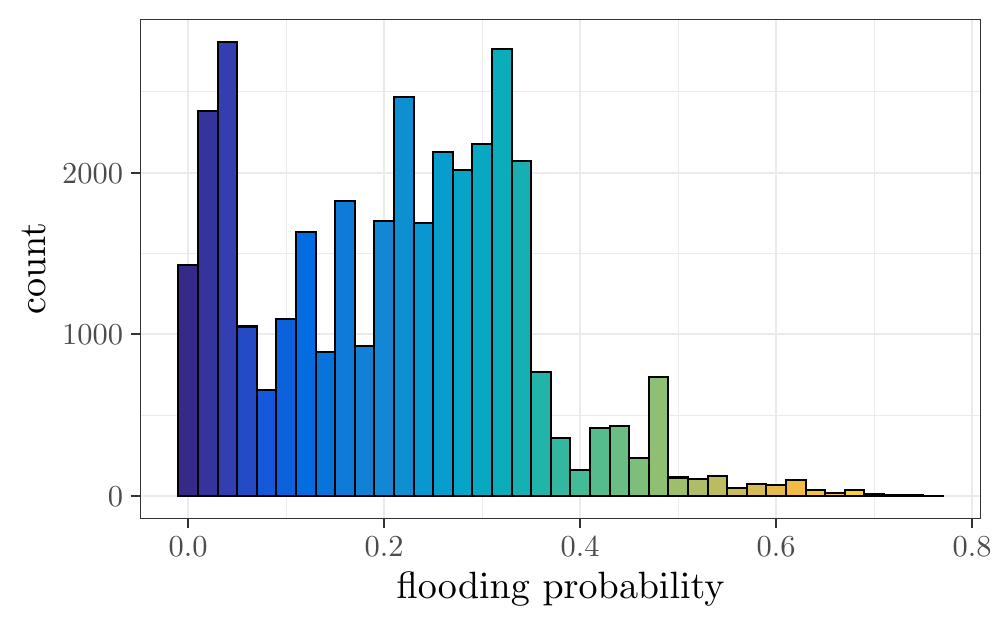}		
	\end{minipage}
	\caption{(left) Empirical flooding probability (EFP) over the 131 map replicates. (right) Histogram of the EFP. Both panels only consider nonzero values of the EFP. For the EFP, the contour line at 0.4 is shown in red.}
	\label{fig:toyExample8BRGMProb}
\end{figure}

\paragraph{Design of experiments (DoE).} The complexity of our model increases with the number of spatial design points $S$. Therefore, we focus the computational budget on a subset of spatial locations. First, we consider locations where the empirical flooding probability (EFP) over the 131 flood scenarios is nonzero. This results in a total of $34\times 10^3$ (approximately) possible candidates {where $\EFP \in (0, 0.8]$} (Figure \ref{fig:toyExample8BRGMProb}, left). {Second, we conveniently divide data into two classes. The DoE for the first class targets the neighborhood of the road network. The  DoE of the second domain is space-filling over the entire domain. After analysis of the flood maps using expert knowledge, we noticed that spatial locations associated with the main road network led to $\operatorname{EFP} \in [0.4, 0.8]$. Thus, we define the second class with locations where $\operatorname{EFP} \in (0, 0.4)$.} For each class, a k-means clustering scheme is applied \citep[see, e.g.,][]{Hartigan1979Kmeans}, where the closest point to each cluster will be part of the DoE. The clustering scheme uses the spatial coordinates and the EFP as inputs, i.e., $\Bx_{\text{clustering}} = (x_1,x_2,\EFP)$. The influence of the EFP contributes to grouping spatially close points into different clusters if they exhibit different flooding probabilities. Note that the number of clusters $\kappa_1, \kappa_2$ are hyperparameters to be defined depending on how many spatial design points we consider per class. {While $\kappa_1$ defines the desired number of design points with significant EFP values in the resulting DoE, $\kappa_2$ will control the number of design points used for space filling.} We also consider three locations of interest from the neighboring district: the town hall, the gym and the lowest point of the sports field. They will be represented by a square, a circle and a triangle, respectively.

{We must note that alternative approaches for the construction of DoEs, such as the PCA-based approach proposed by \citet{Zhang2008PCACM}, could be considered. However, our aim here is not to provide optimal DoEs for further implementations but to provide DoEs that contain spatial locations where floods have a strong impact while promoting space filling.}

\subsubsection{Leave-one-out (LOO) test}
\label{sec:BRGMresults:subsec:LOO}
To validate our framework, we first test it on an LOO experiment where each scenario from the dataset is predicted using data from the other scenarios. This will give us an idea about the capability of the model for forecasting flood events. To train the 131 models, the same spatial design points are fixed for each scenario in order to apply Kronecker-based computations. For instance, we define a DoE with $S = 103$ locations (including the town hall, the gym and the sports field) using the k-means-based methodology discussed in Section \ref{sec:BRGMresults:subsec:settings}. This leads to $N = 13390$ spatial points that are then used for the covariance parameter estimation. For the selection of the types of kernels used for correlating the functional inputs and spatial locations, different combinations of kernels in Table \ref{tab:fkernels} have been tested. After running the corresponding LOO tests, the use of Mat\'ern 5/2 kernels commonly outperformed any other combination. Therefore, the results here and in further experiments will consider functional and spatial Mat\'ern 5/2 kernels.

\begin{figure}[t!]
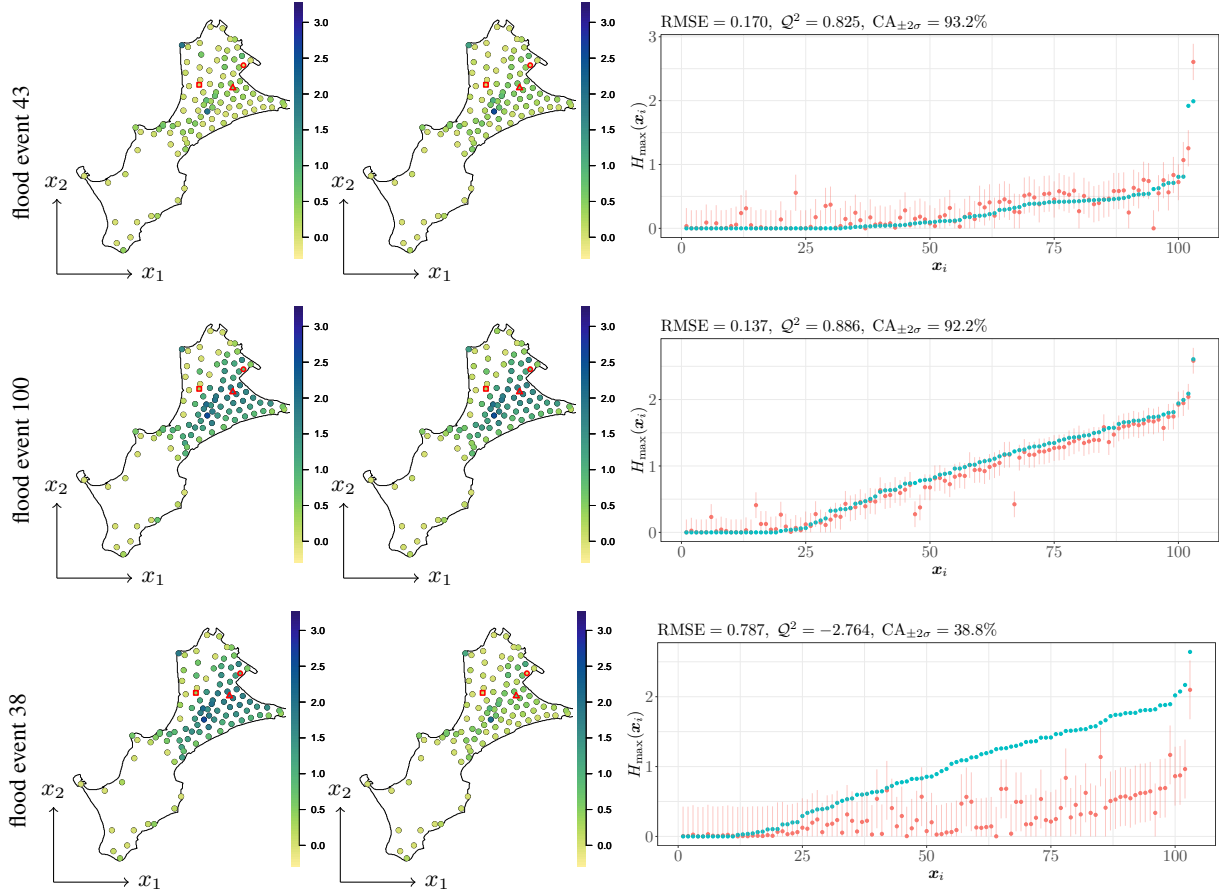

	\centering
	\foreach \i\j in {43,100,38} {
		\begin{minipage}{0.48\textwidth}
			{\footnotesize \rotv{\hskip-7ex flood event \i}}
			\begin{minipage}{0.45\linewidth}
				\begin{tikzpicture}
					\node[anchor=south west,
					xshift=-\textwidth,
					yshift=-\textwidth] (image) at (current page.south west) {
						\includegraphics[width=\textwidth]{toyPaperExample1Gavres103Exp2Trial1MapTest\i.pdf}
					};
					\draw[->] (-\textwidth,-\textwidth)--++(0:1) node[right]{\footnotesize$x_1$};
					\draw[->] (-\textwidth,-\textwidth)--++(90:1) node[above]{\footnotesize$x_2$};
				\end{tikzpicture}
			\end{minipage}
			\hskip1ex
			\begin{minipage}{0.45\linewidth}
				\begin{tikzpicture}
					\node[anchor=south west,
					xshift=-\textwidth,
					yshift=-\textwidth] (image) at (current page.south west) {
						\includegraphics[width=\textwidth]{toyPaperExample1Gavres103Exp2Trial1MapTestPred\i.pdf}
					};
					\draw[->] (-\textwidth,-\textwidth)--++(0:1) node[right]{\footnotesize$x_1$};
					\draw[->] (-\textwidth,-\textwidth)--++(90:1) node[above]{\footnotesize$x_2$};
				\end{tikzpicture}
			\end{minipage}
		\end{minipage}
		\hskip1ex
		\begin{minipage}{0.48\textwidth}
			\vskip1.5ex		
			\includegraphics[width=\textwidth]{toyPaperExample1Gavres103Exp2Trial1IntConfProfile\i.pdf}
		\end{minipage}
	}
	\caption{$H_{\max}$ predictions for the experiment in Section \ref{sec:BRGMresults:subsec:LOO}: (left) ground truth, (middle) predictions, and (right) ground truth (blue) vs. the predictive mean ($\pm$ two-standard deviation confidence intervals, red). For the flood event profiles, the town hall, gym and sports field are represented by a red square, a red circle and a red triangle, respectively. For the right side panels, the $H_{\max}$ values are sorted in increasing order of the true observations. The $\mbox{RMSE}$, $\mathcal{Q}^2$ and $\mbox{CA}_{\pm2\sigma}$ values are shown on top of the right side panels.}
	\label{fig:BRGMnbExp1}
	\vskip-3ex	
\end{figure}

Examples of the LOO predictions are shown in Figure \ref{fig:BRGMnbExp1}. Since negative predictions do not have physical meaning in our application, we set them to zero for further analysis. From scenarios 43 and 100, our framework properly infers their flood levels, leading to accurate RMSE and $\mathcal{Q}^2$ values.\footnote{The $\mathcal{Q}^2$ in \eqref{eq:Q2} is invalid when the variance of the test data (the flood event that is predicted) is equal to zero. In our application, since some of the scenarios are not flooded, we redefine the $\mathcal{Q}^2$ criterion by normalizing the MSE using the variance of the 131 scenarios instead but considering only spatial locations where the EFP is nonzero.} Note that for both scenarios, $\mbox{CA}_{\pm 2\sigma}$ covers more than 92\% of the test data.

\begin{figure}[t!]
	\centering
	\includegraphics[width=0.7\columnwidth]{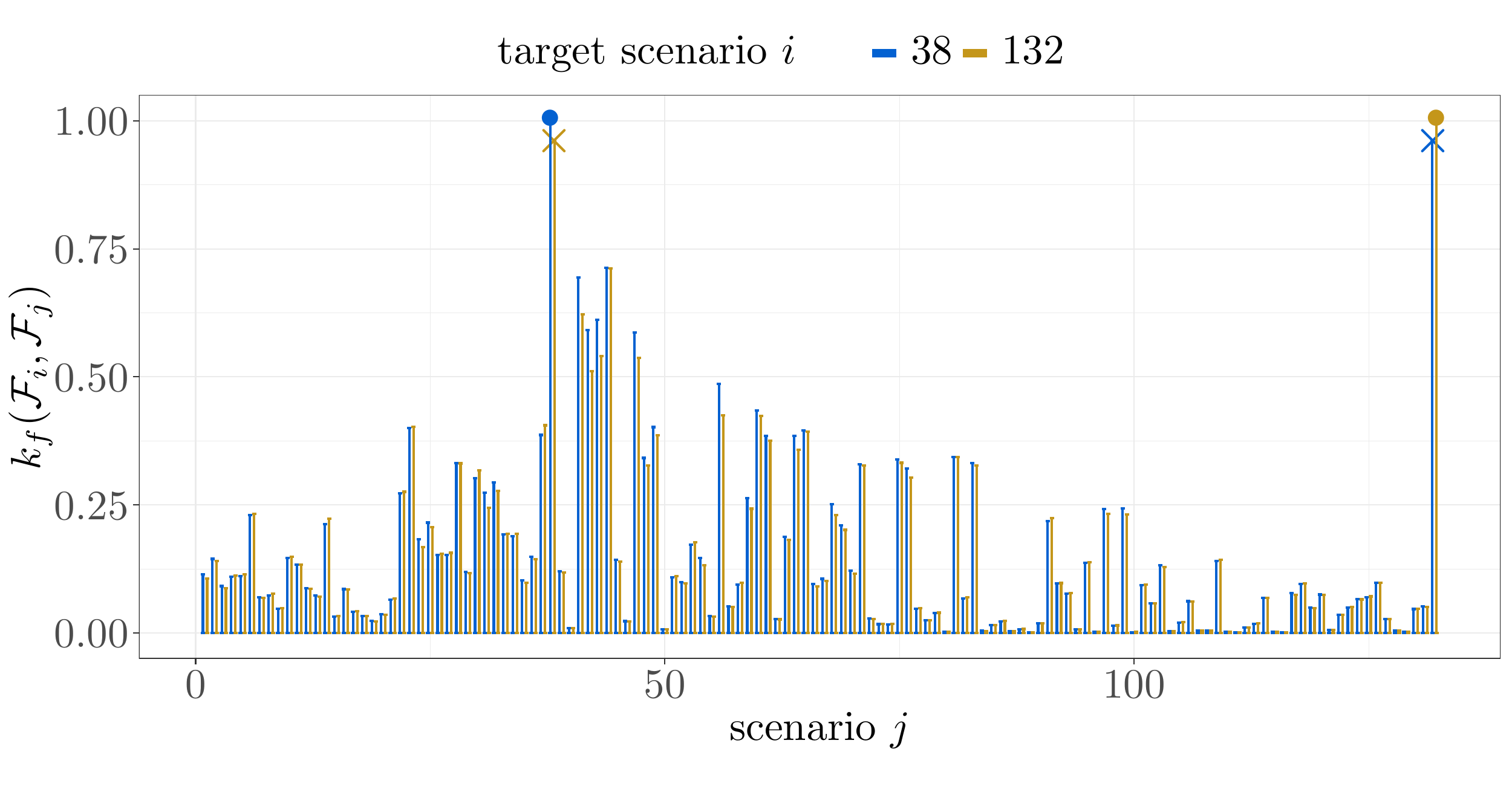} \\
	
	\foreach \i in {38,132} {
		\begin{minipage}{0.485\textwidth}
			\centering
			\includegraphics[width=\textwidth]{toyPaperExample1bGavres103Exp2IntConfProfile\i.pdf} \vskip-3ex
			{\scriptsize flood event \i}
		\end{minipage}
	}
	\caption{(top) Correlations between the hydrometeorological functional conditions of the predicted scenario and those of the learning set. (bottom) The $H_{\max}$ values (sorted in increasing order of the true observations) are shown after adding scenario 132. The panel description is the same as that of Figure \ref{fig:BRGMnbExp1}.}
	\label{fig:BRGMnbExp1Mod}
\end{figure}

Considering the predictions for the 131 LOO tests, they resulted in $\mathcal{Q}^2$ and $\mbox{CA}_{\pm 2\sigma}$ median values of approximately 95.8\% and 99\%, respectively. However, for some scenarios, our framework led to misprediction, as observed for the 38th scenario (see Figure \ref{fig:BRGMnbExp1}). Indeed, in the coastal flooding dataset, there are scenarios that strongly differ from each other or that are unique in their type, and misprediction arises when one of those scenarios is analyzed and if there are no similar flood events available in the learning dataset. As we show in Figure \ref{fig:BRGMnbExp1Mod}, this drawback can be mitigated by incorporating additional and similar scenarios in the learning set of flood events. There, a new flood event, denoted as scenario 132, was added to the LOO test. The hydrometeorological conditions for this scenario are generated by slightly modifying the ($\Tp$, $\Hs_x$, $\Hs_y$, $\U_x$, and $\U_y$) from the 38th scenario but preserving the same profiles for ($\msl$, $\tide$, and $\surge$). From Figure \ref{fig:BRGMnbExp1Mod}, the correlations between the hydrometeorological functional inputs of scenarios 38 and 132 are stronger than if we compare their inputs with those of other scenarios. Note also that predictions for the 38th scenario were significantly improved by exploiting data from the 132nd scenario, and vice versa.

\subsubsection{Influence of the number of learning scenarios}
\label{sec:BRGMresults:subsec:nbMaps}
As stated in Section \ref{sec:BRGMresults:subsec:LOO}, the performance of the model depends on the availability and diversity of learning flood scenarios. In this experiment, we stress the impact of enriching the set of learning scenarios for forecasting purposes. We focus on the prediction of (21) historical flood and no-flood events and (16) slightly reinforced historical flood events, i.e., the first 37 scenarios of the dataset. Those scenarios are predicted using data from the remaining 94 simulated scenarios. Since the latter set of scenarios was generated to cover a wide range of hydrometeorological conditions (see Section \ref{sec:BRGMapp}), accurate predictions are also expected when considering only a few learning scenarios.

We train models considering different subsets of the 94 learning scenarios. The selection of those scenarios is performed via k-means clustering using the scalar representations of the hydrometeorological forcing conditions proposed in \citep{Rohmer2020Nuanced} as inputs. A k-means algorithm considering 20 clusters is applied where the index (number of scenarios) of the closest point (set of scalar hydrometeorological conditions) to each cluster is retained to construct the learning set of scenarios. This leads to an initial set of $R_o = 20$ learning scenarios. To enrich the learning dataset, new scenarios are added by repeating the same procedure but applying the k-means algorithm to the scenarios that have not been previously chosen. Therefore, for each addition, we have that $R_i = R_{i-1} + \Delta R_i$, for $i = 1, \ldots, N_R$, where $\Delta R_i$ is the number of added scenarios and $N_R$ is the number of enrichments. After defining the learning sets of flood scenarios, we then fit the corresponding GP models for the resulting 37 test cases. Both the training and prediction steps are performed considering the spatial design points proposed in Section \ref{sec:BRGMresults:subsec:LOO}.

\begin{figure}
	\centering
	\includegraphics[width=0.9\columnwidth]{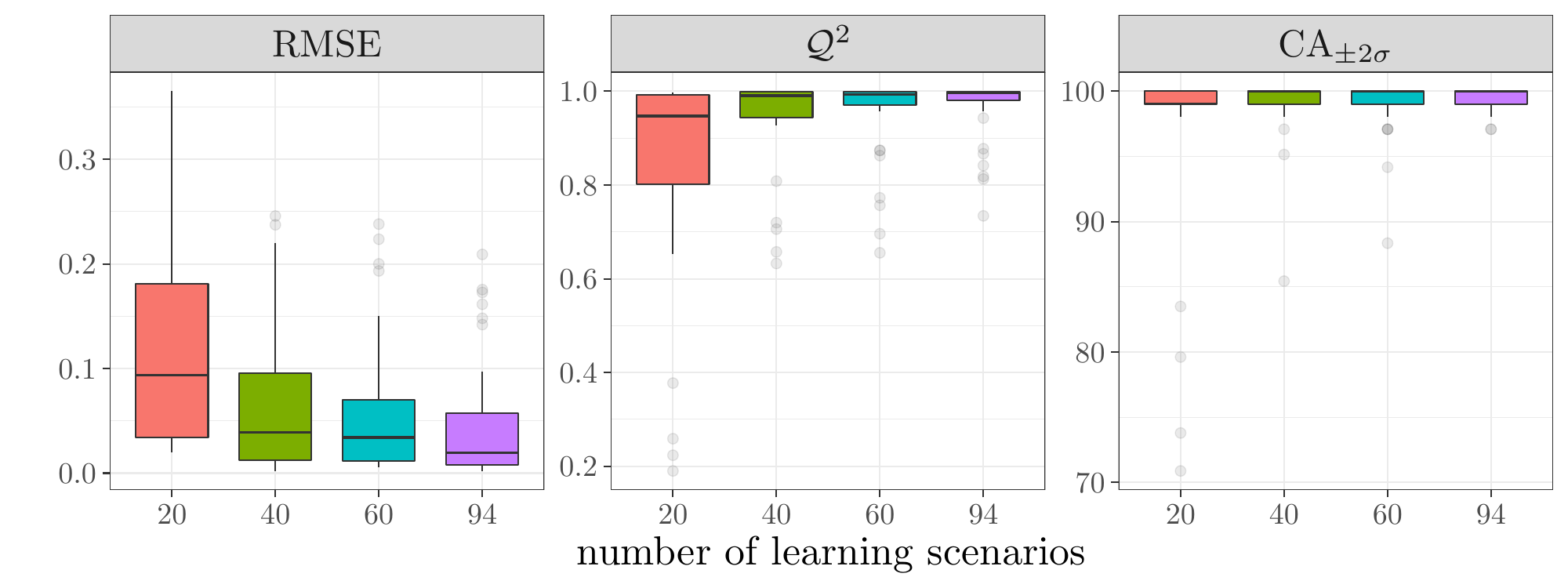}
	\caption{RMSE (left), $\mathcal{Q}^2$ (middle) and $\mbox{CA}_{\pm 2\sigma}$ (right) results for the experiment in Section \ref{sec:BRGMresults:subsec:nbMaps}. The boxplots are computed over the first 37 scenarios of the dataset in Section \ref{sec:BRGMapp}. The results are shown for different numbers of learning scenarios. The learning scenarios are taken only from the remaining 94 simulated scenarios of the coastal flooding dataset.}
	\label{fig:BRGMnbExp1FullMapsErrors37Maps}
\end{figure}

\begin{figure}[t!]
	\centering
	\includegraphics[width=\columnwidth]{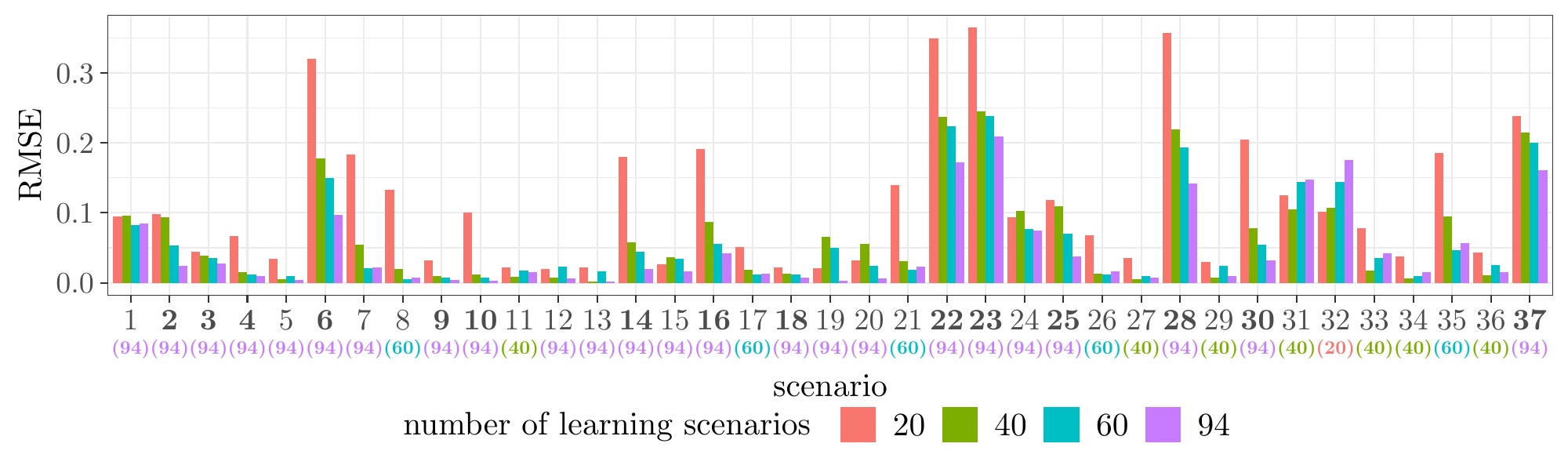}
	\caption{RMSE results of predictions for the first 37 scenarios of the dataset in Section \ref{sec:BRGMapp} and considering $R = 20, 40, 60, 94$. The bars represent the magnitude of the RMSE criterion for each scenario. The fifteen scenarios where the RMSEs improved at each of the three increases of $R$ are in bold. The subindices at the bottom of the x-axis designate which case provided the smaller RMSE.}
	\label{fig:BRGMnbExp1b}
\end{figure}

Boxplots of the RMSE, $\mathcal{Q}^2$ and $\mbox{CA}_{\pm 2\sigma}$ values of the 37 predictions and considering $R = 20, 40, 60, 94$ learning scenarios are shown in Figure \ref{fig:BRGMnbExp1FullMapsErrors37Maps}. Improvements are obtained on the three performance indicators as $R$ increases, leading to smaller dispersion of the boxplots. For each test scenario, the $\mbox{CA}_{\pm 2\sigma}$ remains satisfactory regardless of $R$, and the RMSE and $\mathcal{Q}^2$ results are commonly outperformed when considering a larger learning dataset. More precisely, 24 of the 37 predictions are improved by considering all 94 learning scenarios. However, for some cases (see Figure \ref{fig:BRGMnbExp1b}, e.g., the RMSE results for scenarios 31 and 32), the prediction quality decreased.

\begin{figure}[t!]
	\centering
	\subfigure[\label{fig:BRGMnbExp1b32:subfig:correlations}]{\includegraphics[height=0.26\columnwidth]{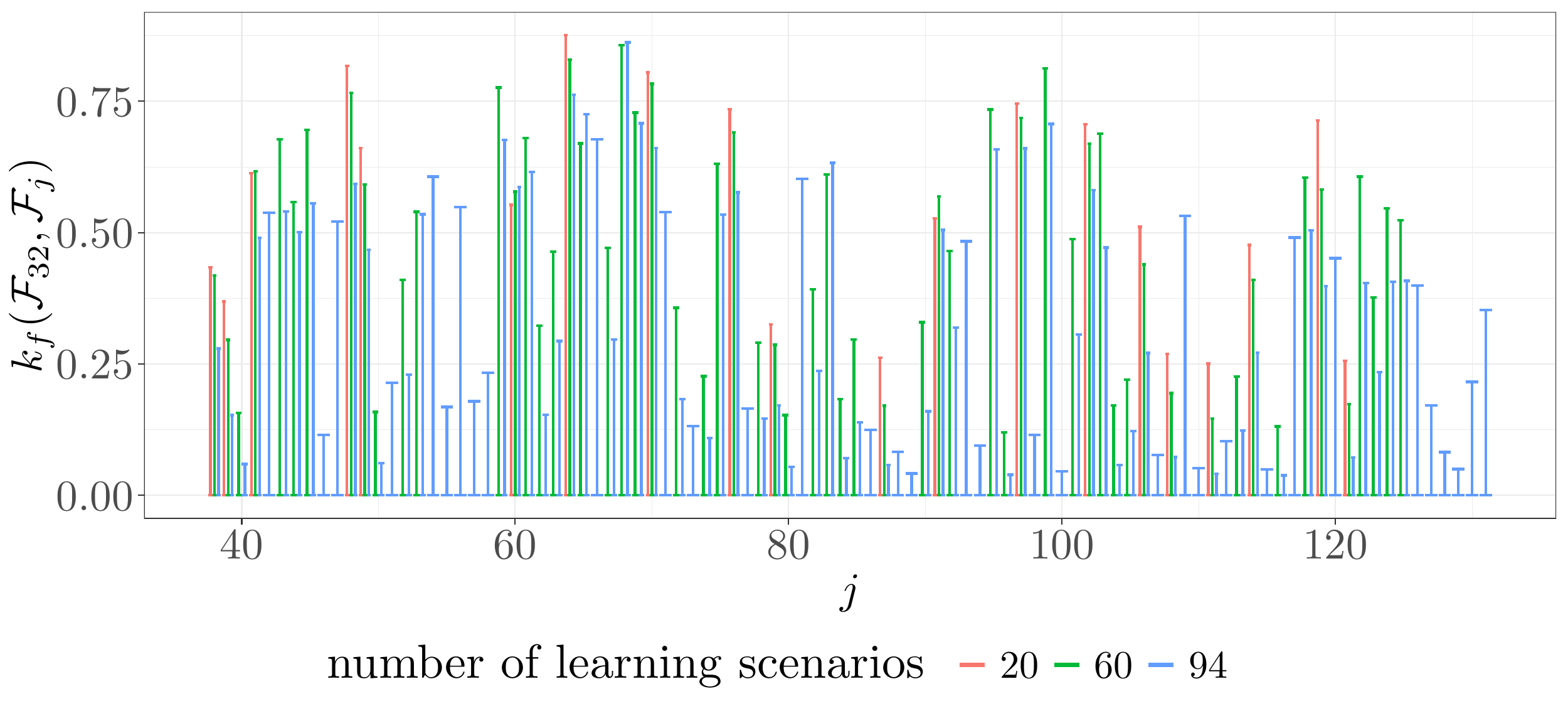}}
	\subfigure[\label{fig:BRGMnbExp1b32:subfig:finputs}]{\includegraphics[height=0.26\columnwidth]{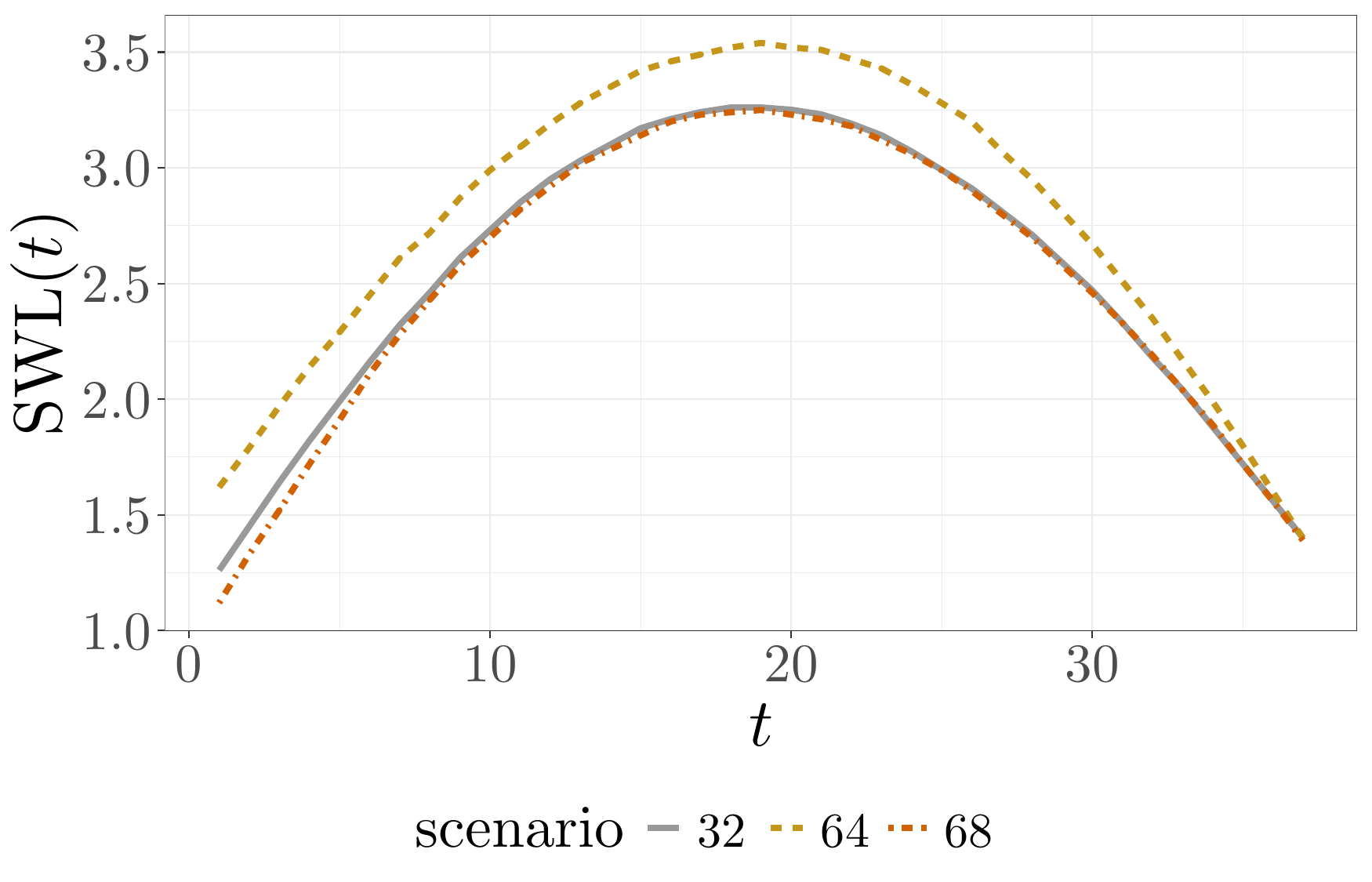}}
	
	\subfigure[\label{fig:BRGMnbExp1b32:subfig:maps}]{
		\foreach \i in {32,64,68}{
			\begin{minipage}{0.3\textwidth}
				\centering 				
				{\footnotesize flood event \i} \\				
				\begin{tikzpicture}
					\node[anchor=south west,
					xshift=-\textwidth,
					yshift=-\textwidth] (image) at (current page.south west) {
						\includegraphics[width=\textwidth]{toyPaperExample1Gavres10332Map\i-1.png}
					};
					\draw[->] (-\textwidth,-\textwidth)--++(0:1) node[right]{\footnotesize$x_1$};
					\draw[->] (-\textwidth,-\textwidth)--++(90:1) node[above]{\footnotesize$x_2$};
				\end{tikzpicture}
			\end{minipage}
		}
	}
	\caption{\subref{fig:BRGMnbExp1b32:subfig:correlations} Correlations between the hydrometeorological conditions of scenario 32 and those of the learning set. The results considering different numbers of learning scenarios are shown. \subref{fig:BRGMnbExp1b32:subfig:finputs} Time series of the still water level ($\swl(t) = \msl(t) + \tide(t) + \surge(t)$) for scenarios 32, 64 and 68. \subref{fig:BRGMnbExp1b32:subfig:maps} True $H_\max$ maps of the scenarios under analysis.}
	\label{fig:BRGMnbExp1b32}
\end{figure}

We first analyze results for the 32nd scenario. By comparing the similarity between the hydrometeorological conditions using either 20 or 94 learning scenarios, the strongest correlations are provided with scenarios 64 and 68, respectively (Figure \ref{fig:BRGMnbExp1b32:subfig:correlations}). For the ease of the discussion, we focus on the comparison of the still water level (SWL), which is given by $\swl(t) = \msl(t) + \tide(t) + \surge(t)$.\footnote{By performing sensitivity analysis (SA) based on GPs, models were more sensitive to the $\msl$, $\tide$ and $\surge$ rather than the other hydrometeorological drivers. For the SA, the \texttt{sensitivity} package \citep{Bertrand2020Sensitivity} was adapted to account for functional data.} Figure \ref{fig:BRGMnbExp1b32:subfig:finputs} shows that the SWL of scenario 68 is actually closer to that of scenario 32. Therefore, after adding scenario 68 to the learning dataset (event already added for $R = 60$), models may consider stronger correlations with respect to this scenario rather than the 64th scenario. However, while scenarios 32 and 64 correspond to moderate flood events, the 68th scenario corresponds to a weaker flood event (Figure \ref{fig:BRGMnbExp1b32:subfig:maps}), explaining the misprediction when considering such scenarios in the learning dataset.

Similar to scenario 32, similar behaviors were observed in other scenarios where the RMSE systematically decreased as $R$ increased. To avoid this type of drawback, we can enrich the learning dataset with additional flood scenarios and/or consider adding complementary hydrometeorological conditions (or prior information on flood events) as inputs of the GP model that may help to discriminate scenarios. Figure \ref{fig:BRGMnbExp1b32:subfig:finputs} shows that small variations in the hydrometeorological conditions may lead to events with and without floods. Indeed, this remind us that floods occur only if the instantaneous water level resulting from the joint action of the MSL, tide, surge, waves and wind exceeds the level of the (natural or man-made) coastal defense crests. In that case, our framework must instead be adapted to learn the critical forcing conditions leading the instantaneous water level to exceed the defense crests.

\subsubsection{Influence of the number of spatial design points}
\label{sec:BRGMresults:subsec:nbPoints}
\begin{figure}[t!]
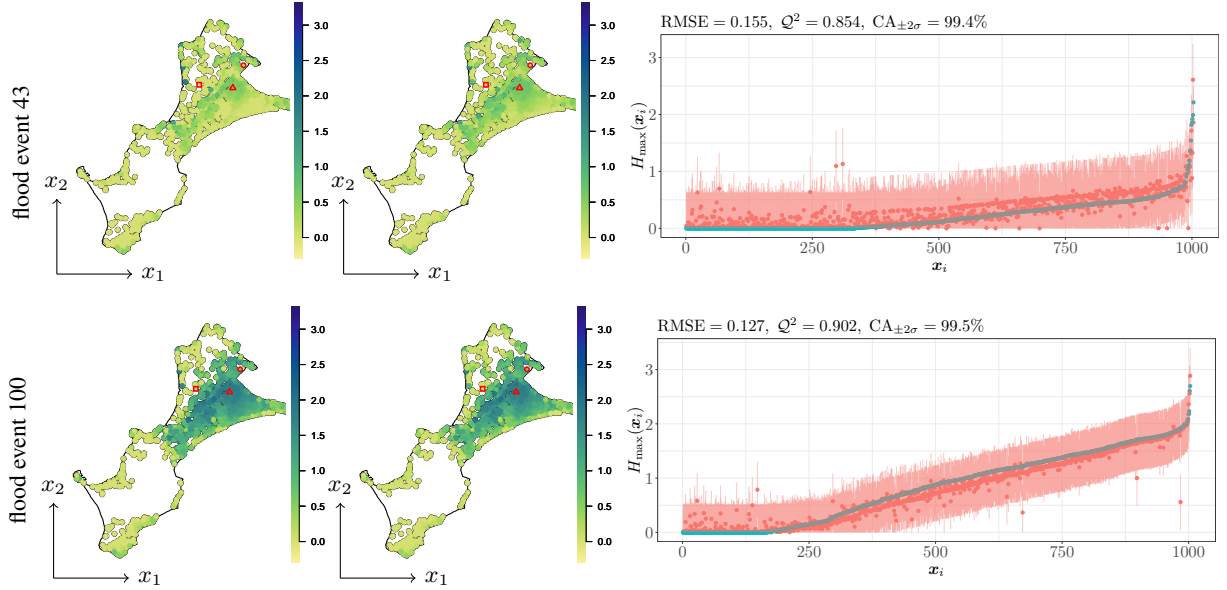

	\centering
	\foreach \i in {43,100} {
		\begin{minipage}{0.48\textwidth}
			{\footnotesize \rotv{\hskip-7ex flood event \i}}
			\begin{minipage}{0.45\linewidth}
				\begin{tikzpicture}
					\node[anchor=south west,
					xshift=-\textwidth,
					yshift=-\textwidth] (image) at (current page.south west) {
						\includegraphics[width=\textwidth]{toyPaperExample5Gavres1003Exp2Trial1MapTest\i.pdf}
					};
					\draw[->] (-\textwidth,-\textwidth)--++(0:1) node[right]{\footnotesize$x_1$};
					\draw[->] (-\textwidth,-\textwidth)--++(90:1) node[above]{\footnotesize$x_2$};
				\end{tikzpicture}
			\end{minipage}
			\hskip1ex
			\begin{minipage}{0.45\linewidth}
				\begin{tikzpicture}
					\node[anchor=south west,
					xshift=-\textwidth,
					yshift=-\textwidth] (image) at (current page.south west) {
						\includegraphics[width=\textwidth]{toyPaperExample5Gavres1003Exp2Trial1MapTestPred\i.pdf}
					};
					\draw[->] (-\textwidth,-\textwidth)--++(0:1) node[right]{\footnotesize$x_1$};
					\draw[->] (-\textwidth,-\textwidth)--++(90:1) node[above]{\footnotesize$x_2$};
				\end{tikzpicture}
			\end{minipage}
		\end{minipage}
		\hskip1ex
		\begin{minipage}{0.48\textwidth}
			\vskip1.5ex		
			\includegraphics[width=\textwidth]{toyPaperExample5Gavres1003Exp2Trial1IntConfProfile\i.pdf}
		\end{minipage}
	}
	\caption{The $H_{\max}$ predictions for the experiments in Figure \ref{fig:BRGMnbExp1} considering 1003 spatial design points per flood scenario. The panel description is the same as that in Figure \ref{fig:BRGMnbExp1}.}
	\label{fig:BRGMnbExp2}
\end{figure}

We now assess the impact of the number of spatial design points $S$ in predictions. To do so, we repeat the experiment in Section \ref{sec:BRGMresults:subsec:LOO} but consider 900 additional design points per flood event, i.e., $S=1003$. This leads to a total of $N \sim 1.3 \times 10^{5}$ spatial design points. The same parameterization proposed in Section \ref{sec:BRGMresults:subsec:LOO} is used here. Figure \ref{fig:BRGMnbExp2} shows the predictions for scenarios 43 and 100. As also observed in Figure \ref{fig:BRGMnbExp1}, the model provides accurate predictions, leading to $\mathcal{Q}^2$ and $\mbox{CA}_{\pm 2\sigma}$ values above 85\% and 99\%, respectively.
\begin{figure}[t!]
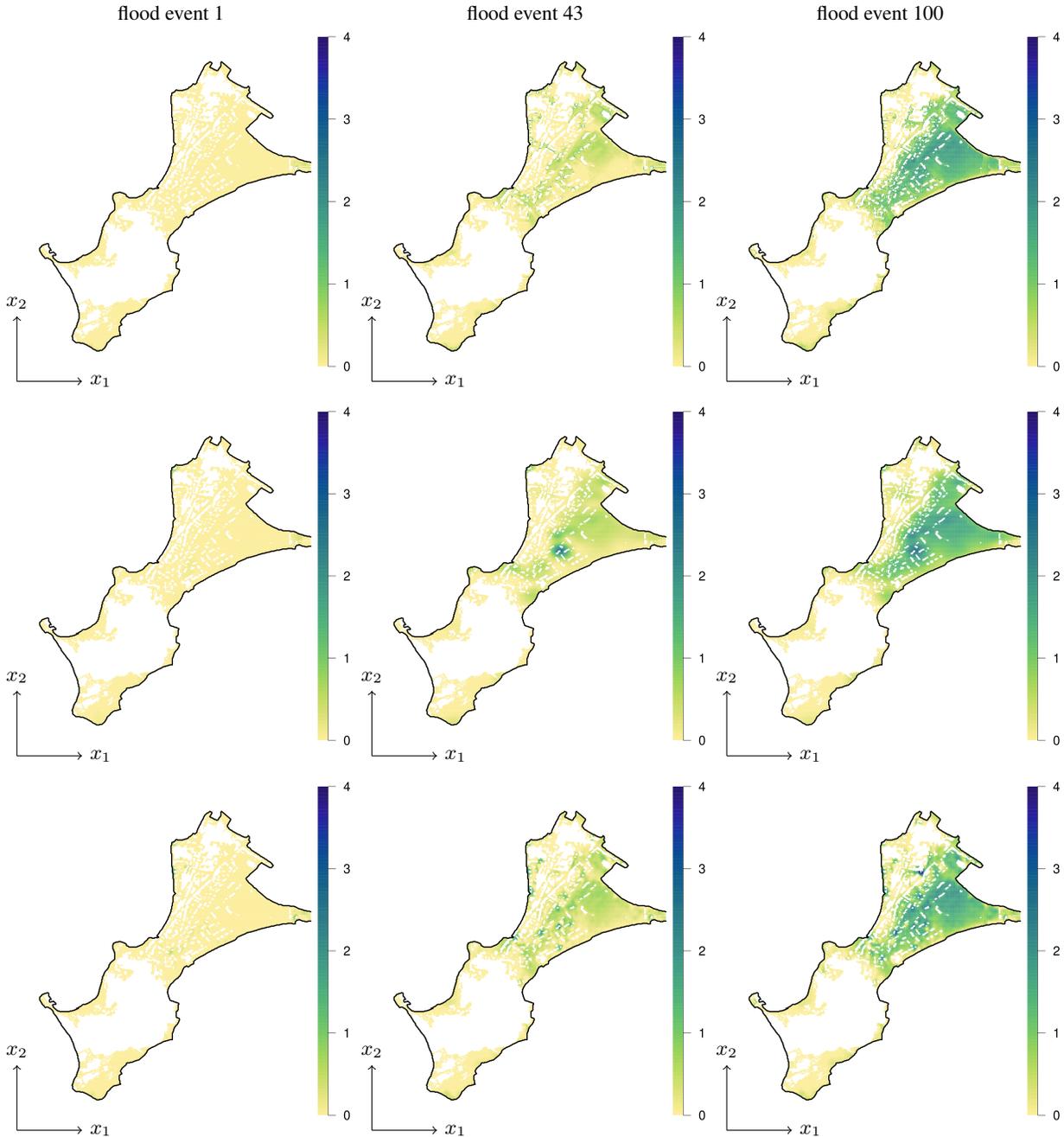

	\centering
	\foreach \i in {1,43,100} {
		\begin{minipage}{0.32\textwidth}
			\centering 
			{\footnotesize flood event \i} \\
			\begin{tikzpicture}
				\node[anchor=south west,
				xshift=-\textwidth,
				yshift=-\textwidth] (image) at (current page.south west) {
					\includegraphics[width=\textwidth]{toyPaperExample5Gavres1003MapNonZero\i-1.png}
				};
				\draw[->] (-\textwidth,-\textwidth)--++(0:1) node[right]{\footnotesize$x_1$};
				\draw[->] (-\textwidth,-\textwidth)--++(90:1) node[above]{\footnotesize$x_2$};
			\end{tikzpicture}
			\begin{tikzpicture}
				\node[anchor=south west,
				xshift=-\textwidth,
				yshift=-\textwidth] (image) at (current page.south west) {
					\includegraphics[width=\textwidth]{toyPaperExample1Gavres103Exp2Trial1MapPred\i-1.png}
				};
				\draw[->] (-\textwidth,-\textwidth)--++(0:1) node[right]{\footnotesize$x_1$};
				\draw[->] (-\textwidth,-\textwidth)--++(90:1) node[above]{\footnotesize$x_2$};
			\end{tikzpicture}
			\begin{tikzpicture}
				\node[anchor=south west,
				xshift=-\textwidth,
				yshift=-\textwidth] (image) at (current page.south west) {
					\includegraphics[width=\textwidth]{toyPaperExample5Gavres1003Exp2Trial1MapPred\i-1.png}
				};
				\draw[->] (-\textwidth,-\textwidth)--++(0:1) node[right]{\footnotesize$x_1$};
				\draw[->] (-\textwidth,-\textwidth)--++(90:1) node[above]{\footnotesize$x_2$};
			\end{tikzpicture}
		\end{minipage}
	}
	\caption{Predictions for the experiment in Section \ref{sec:BRGMresults:subsec:LOO} considering only spatial locations with nonzero EFPs. The panels show the true $H_{\max}$ maps (top) and the predictions using either 103 (middle) or 1003 design points per flood event (bottom).}
	\label{fig:BRGMnbExp1FullMaps}
\end{figure}
{In practice, we may be interested in predicting flood areas rather than locally predicting values only at the subset of $S$ spatial locations. This can be achieved by applying the conditional predictive GP formulas to the spatial domain of interest (see the discussion in Sections \ref{sec:functionalGPs} and \ref{sec:spatGPs}). Predictions of scenarios 1, 43 and 100 considering only spatial locations with nonzero EFPs ($\sim 34\times 10^3$ points) are shown in Figure \ref{fig:BRGMnbExp1FullMaps} for $S = 103, 1003$. For both cases, our framework tends to capture the intensity of those flood events. While spatial profiles are smoother by considering $S = 103$, a better predictability resolution is obtained for $S = 1003$. This is caused by the difference in the magnitude of the estimated length-scales $(\ell_{x,1},\ell_{x,2})$ [m]. This should remind us that for larger values of $(\ell_{x,1},\ell_{x,2})$, models lead to stronger spatial correlations and therefore to smoother predictions. Here, while the former GP model with $S=103$ resulted in length-scale (median) estimations equal to $\hat{\ell}_{x,1,103} = 42.7$ and $\hat{\ell}_{x,2,103} = 135.1$, the latter model with $S=1003$ resulted in smaller length-scales: $\hat{\ell}_{x,1,1003} = 19.1$ and $\hat{\ell}_{x,2,1003} = 28.4$. For $S = 1003$, overestimations commonly occur in the areas surrounding buildings (small white zones). Indeed, the model overestimates $H_\max$ values at the center of buildings (locations that were never observed in the training step; see the discussion in Section \ref{sec:BRGMresults:subsec:settings}) where those values are actually equal to zero. Therefore, because of the smoothness condition of GPs, predictions in the buildings' neighborhood are affected.
	
	Table \ref{tab:BRGMnbExp2HmaxProp} assesses the quality of the predictions in Figure \ref{fig:BRGMnbExp1FullMaps} according to the flood categories recommended by the French Risk Prevention Plan \citep{PPRL2014}. In this plan, depending on the impact of the flood (e.g., the water height of the neighboring district and/or in the road network), events are classified as follows: minor if $H_\max \leq 0.5$, moderate if $0.5 < H_\max \leq 1$, serious if $1 < H_\max \leq 1.5$ and severe if $H_\max > 1.5$. Then, for scenarios in Figure \ref{fig:BRGMnbExp1FullMaps}, we pointwise assign predictions $\widehat{H}_\max$ to their corresponding categories, and we compare the resulting proportions [\%] per flood category with respect to those led by the true observations $H_\max$. From Table \ref{tab:BRGMnbExp2HmaxProp}, we can observe that both models, considering $S = 103$ or $S = 1003$, globally lead to reliable flood proportions. Misclassification between consecutive categories mainly results from the overestimation or underestimation around the $H_\max$ threshold that defines the limit of each category. As an example, for scenario 43, a significant number of $H_\max$ values close to $0.5$ were assigned in the moderate category when they actually belong to the minor category.}{} 

\begin{table}[t!]
	\centering
	\caption{Quality assessment of predictions in Figure \ref{fig:BRGMnbExp1FullMaps} considering the flood categories recommended by the French Risk Prevention Plan. Flood events are classified as minor ($H_\max \leq 0.5$), moderate ($0.5 < H_\max \leq 1$), serious $(1 < H_\max \leq 1.5)$ and severe $(H_\max > 1.5)$. The proportions [\%] per category are computed using spatial locations with nonzero EFPs. The results are shown considering the true observations ($H_\max := H$) and predictions using either $S_1 = 103$ ($\widehat{H}_{S_1}$) or $S_2 = 1003$ ($\widehat{H}_{S_2}$). The closest proportions to the ones provided by $H$ (given in gray) are highlighted in bold.}
	\label{tab:BRGMnbExp2HmaxProp}	
	\begin{tabular}{c|ccc|ccc|ccc}
		\toprule				
		\multirow{2}{*}{Flood} & \multicolumn{9}{c}{Proportions [\%] per Category} \\		
		\multirow{2}{*}{Category} & \multicolumn{3}{c|}{Scenario 1} & \multicolumn{3}{c|}{Scenario 43}  & \multicolumn{3}{c}{Scenario 100}  \\
		& $H$ & $\widehat{H}_{S_1}$ & $\widehat{H}_{S_2}$ & $H$ & $\widehat{H}_{S_1}$ & $\widehat{H}_{S_2}$ & $H$ & $\widehat{H}_{S_1}$ & $\widehat{H}_{S_2}$ \\
		\midrule
		minor    	& \textbf{\color{gray}99.9} & \textbf{99.7} & 99.6 & \textbf{\color{gray}90.9} & \textbf{86.6} & 83.3 & \textbf{\color{gray}50.6} & 51.3 & \textbf{50.8} \\
		moderate    & \textbf{\color{gray}0.1}  & \textbf{0.3}  & \textbf{0.3}  & \textbf{\color{gray}7.9}  & \textbf{12.0} & 15.4 & \textbf{\color{gray}16.1} & \textbf{17.3} & 19.9 \\
		serious    & \textbf{\color{gray}0.0}  & \textbf{0.0}  & 0.1  & \textbf{\color{gray}0.9}  & 0.8  & \textbf{0.9}  & \textbf{\color{gray}20.0} & \textbf{20.9} & 18.0 \\
		severe & \textbf{\color{gray}0.0}  & \textbf{0.0}  & \textbf{0.0}  & \textbf{\color{gray}0.3}  & 0.6  & \textbf{0.4}  & \textbf{\color{gray}13.4} & 10.5 & \textbf{11.2} \\
		\bottomrule	
	\end{tabular}
\end{table}

\begin{figure}
	\centering
	\includegraphics[width=\columnwidth]{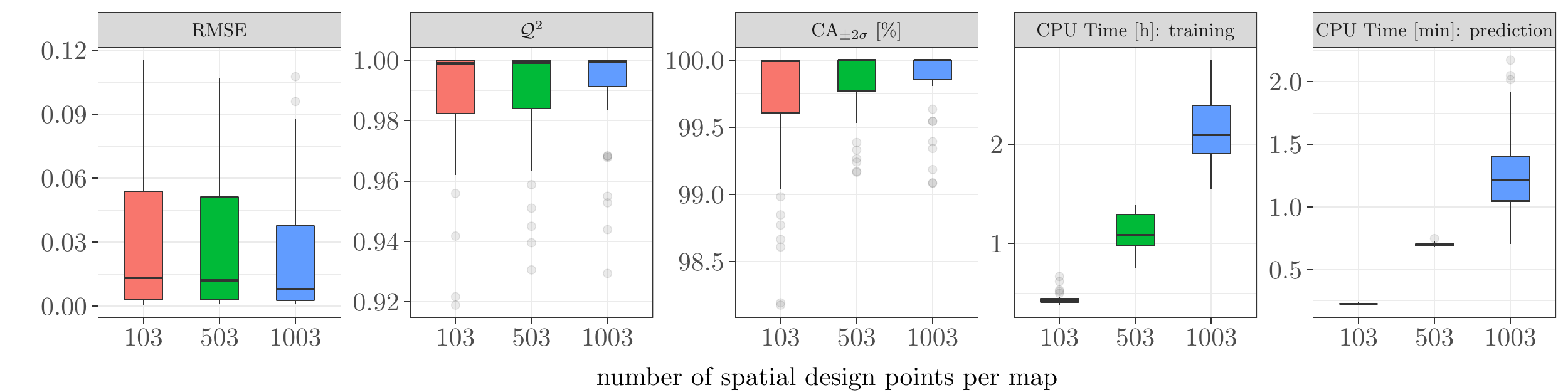}	
	\caption{Performance indicators for the experiment in Section \ref{sec:BRGMresults:subsec:nbPoints} considering the first 37 scenarios of the dataset in Section \ref{sec:BRGMapp}. The results are shown for $S = 103, 503, 1003$.}
	\label{fig:BRGMnbExp1FullMapsErrors}
\end{figure}

Finally, Figure \ref{fig:BRGMnbExp1FullMapsErrors} shows the boxplots of the performance indicator values for the LOO predictions and considering $S = 103, 503, 1003$. As discussed in Sections \ref{sec:BRGMresults:subsec:LOO} and \ref{sec:BRGMresults:subsec:nbMaps}, to avoid misprediction due to the uniqueness of some scenarios, the analysis is focused on the first 37 flood events. The indicators (RMSE, $\mathcal{Q}^2$, and $\mbox{CA}_{\pm 2\sigma}$) are computed considering only spatial locations with nonzero EFPs (see Figure \ref{fig:toyExample8BRGMProb}). They improve as $S$ increases, leading to smaller dispersion of the boxplots. In terms of the computational costs, note that although the CPU times for both training and prediction steps increase as $S$ increases, they remain tractable.\footnote{These experiments were executed on a single core of an HP cluster with an AMD quadprocessor and 256 GB RAM.} In practice, since the training step is executed only once and can be computed offline, we only need to be aware of the computational costs of the prediction step. Here, the prediction of a single map takes a couple of minutes, an advantage compared to the couple of days required by numerical simulators. Hence, this makes possible the use of our framework for FEWSs.

\section{Conclusions}
\label{sec:conclusions}
{In this paper, we established a GP-based surrogate model that accounts for the effect of time-varying forcing conditions (e.g., tides, storms, surges, etc.) and provides spatially varying inland flood information (maximal inland water height, $H_\max$).} We demonstrated that our framework can be easily applied without any parameterization of the hydrometeorological drivers in scalar representations, a step that is not always simple and that can lead to misleading predictions. To the best of our knowledge, our proposal is the first to consider both inputs and outputs of a coastal computer code as functions.

The proposed GP model was built on the proposition of a separable kernel that correlates both hydrometeorological forcing conditions (functional inputs) and spatial input locations. The resulting process can be seen as a multioutput GP where correlated spatial outputs are driven by multiple hydrometeorological functional inputs. Efficient implementations were explored by considering Kronecker-structured operations and/or sparse-variational approximations. This led to models that run fast for up to hundreds of outputs and thousands of spatial design points. Both Python and R codes were provided.

Our framework was widely tested on various examples considering different situations depending on data availability. According to the numerical experiments, our approach led to accurate predictions that outperformed standard GP implementations that cannot account for functional data. For forecasting unobserved spatial events, we numerically showed the convergence of our approach as the number of learning flood events increased. We demonstrated that accurate predictions can be obtained within tractable time lapses in the coastal flooding application. More precisely, predictions were obtained on the order of minutes compared to the couple of days required by dedicated simulators. This is key for FEWSs. We must note that the performance of the proposed framework depends on the availability and diversity of learning flood scenarios. The richer and diverse the learning dataset is, the better the predictability.

The framework presented here can be improved in different ways. A natural extension relies on projecting the spatial outputs onto (truncated) basis representations such as wavelets. In that case, functional inputs can be treated as discussed in this paper, and Gaussian priors will be placed over the space of the coefficients of representations rather than the output space. This would lead to significant improvements in terms of the computational costs for applications with large learning datasets, e.g., involving thousands of outputs and tens of thousands of design points. The nature of coastal flooding data, more precisely, the nonnegative, zero-inflated and discontinuous patterns (as illustrated by Figure \ref{fig:toyExample8BRGMMaps}), cannot be easily learned using purely GP assumptions. Hence, alternative (nonstationary) GP-based priors may be explored to obtain more realistic surrogate models.

\section*{Acknowledgement}
This work was funded by the ANR (French National Research Agency) under the RISCOPE project (ANR-16-CE04-0011): \href{https://perso.math.univ-toulouse.fr/riscope/}{https://perso.math.univ-toulouse.fr/riscope/}. We thank J. Betancourt (IMT) for his helpful recommendations and for providing 94 of the 131 hydrometeorological scenarios. We also thank F. Gamboa (IMT, ANITI) and T. Klein (IMT, ENAC) for their fruitful discussions within the RISCOPE project. {Part of this research was conducted when AFLL was a postdoctoral researcher at IMT and BRGM.}

\appendix

\section{Projection of Functional Inputs onto PCA Basis Functions}
\label{app:projPCA}
Consider the $\tau$-length functional vector $\Bf = [f(t_1), \ldots, f(t_\tau)]^\top$. Let $\BF = [\Bf_{1}, \ldots, \Bf_{N}]^\top$ be a matrix containing $N$ replicates of $\Bf$. The PCA of $\BF$ can be obtained using the variance-covariance matrix \citep{Ramsay2005functional}. Denote the matrix $\BF_{c}$ as the centered version of $\BF$, where the mean of each column of $\BF$ is subtracted from the column. Then, the variance-covariance matrix is given by
\begin{equation}
	\BOmega_\tau = \frac{1}{N} \BF_{c}^{\top} \BF_{c} = \sum_{j=1}^{\tau} \lambda_{j} \Bnu_{j} \Bnu_{j}^\top,
	\label{eq:eig}
\end{equation}
where $\lambda_{1} \geq \cdots \geq \lambda_{\tau}$ and $\Bnu_{1}, \ldots, \Bnu_{\tau}$ are the corresponding eigenvalues and eigenvectors of $\BOmega_\tau$, respectively. In practice, \eqref{eq:eig} is commonly truncated after reaching a predefined inertia $\mathcal{I}_\ast$, where the contribution of each eigencomponent is given by
\begin{equation*}
	\mathcal{I}_{j} = \frac{\lambda_{j}}{\sum_{\kappa = 1}^{\tau} \lambda_{\kappa}}. 
\end{equation*}
Then, the ``optimal'' number of PCA components is obtained by the smallest value of $\tau_\ast$ such that $\sum_{j=1}^{\tau_\ast} \mathcal{I}_{j} \geq \mathcal{I}_\ast$, where $\tau_\ast \leq \tau$. Finally, the eigenvectors of the truncated matrix $\BOmega_{\tau_\ast}$ can be used as basis functions in the linear approximation in \eqref{eq:basisConstr}, i.e., $[\phi_{i,1}, \ldots, \phi_{i,p_i}] = [\Bnu_{1}, \ldots, \Bnu_{\tau_\ast}]$ where $p_i = \tau_\ast$. Due to the orthonormality of $\Bnu_{1}, \ldots, \Bnu_{\tau_\ast}$, the vector of coefficients $\Balpha_{\kappa}$, for $\kappa = 1, \ldots, N$, associated with the replicate $\Bf_{\kappa}$ is given by $\Balpha_{\kappa} = \BPhi^\top \Bf_{\kappa}$ with $\BPhi = [\Bnu_{1}, \ldots, \Bnu_{\tau_\ast}]$.

\section{Kronecker-based Operations for Gaussian Processes with Separable Kernels}
\label{app:kronGPs}
For large datasets, computing the covariance matrix $\BK = (\BK_f \otimes \BK_x) \in \realset{RS \times RS}$ with $R$ outputs and $S$ spatial design points per output may easily run out the memory. This limitation exists when constructing the (lower triangular) Cholesky factorization $\BL = (\BL_f \otimes \BL_x) \in \realset{RS \times RS}$ and its inverse. To mitigate this drawback, instead of computing $\BK$ or $\BL$, we propose solving triangular-structured linear systems involving Kronecker products.

\subsection{Properties of the Kronecker product}
\label{app:kronGPs:subapp:kron}
We first recall the properties of the Kronecker product that are needed in this appendix \citep[see][for further details]{Laub204Kronecker,Alvarez2012kernelReview}. Consider the matrices $\BA, \BA' \in \realset{M \times N}, \BB, \BB' \in \realset{P \times Q}$ and the vectors $\Bu \in  \realset{NQ}, \Bv \in  \realset{MP}$. Then, some useful properties of the Kronecker product are:
\begin{align}
	(\BA \otimes \BB)^\top &= \BA^\top \otimes \BB^\top, \nonumber	\\
	(\BA \otimes \BB)^{-1} &= \BA^{-1} \otimes \BB^{-1}, \label{eq:kron} \\
	(\BA \otimes \BB)(\BA' \otimes \BB') &= (\BA \BA' \otimes \BB \BB'). \nonumber
\end{align}
We are also interested in computing (or solving) linear systems of the form:
\begin{align}
	\Bv = (\BA \otimes \BB) \Bu.
	\label{eq:linSys}
\end{align}
For computing \eqref{eq:linSys}, note that it is possible to rearrange the vectors $\Bu$ and $\Bv$ in matrices in order to avoid constructing $(\BA \otimes \BB)$ (a step that required computing and storing an $MP \times NQ$ matrix). Consider the matrices $\BU \in \realset{Q \times N}$ and $\BV \in \realset{P \times M}$, where the elements of $\Bu$ and $\Bv$ are indexed by columns. Then, \eqref{eq:linSys} is given by
\begin{align}
	\BV = \BB \BU \BA^\top.
	\label{eq:linSysKro}
\end{align}
After computing $\BV$, which involves computing and storing smaller matrices, we can obtain $\Bv$ by vectorizing $\BV$.

\subsection{Computation of the likelihood}
\label{app:kronGPs:subapp:kronLik}
We must note that the complexity of the Gaussian likelihood relies on the computation of the quadratic term:
\begin{align}
	z = \By^\top \BK^{-1} \By,
	\label{eq:quadraticTerm}
\end{align}
with covariance matrix $\BK = \BK_x \otimes \BK_f$ and an observation vector $\By = [y_1, \ldots, y_N]^\top$ with $N = SR$. Note that to achieve numerical simplifications in further steps, we change the order of the Kronecker product proposed in \eqref{eq:ksfproduct} since commonly $R \ll S$. This only implies properly rearranging the indexation of the observations in $\By$, i.e..,
$$\By = [Y(\BfF_1,\Bx_1), \ldots, Y(\BfF_1, \Bx_S), \ldots, Y(\BfF_R, \Bx_1), \ldots, Y(\BfF_R,\Bx_S)]^\top.$$
Then, using the Cholesky factorization of $\BK$, we can show that \eqref{eq:quadraticTerm} is given by
\begin{align*}
	z 
	= {\By}^\top ({\BL} {\BL}^\top)^{-1} {\By}
	= ({\BL}^{-1} {\By})^{\top} ({\BL}^{-1} {\By})
	= {\Ba}^\top {\Ba}.
\end{align*}
For computing $\Ba = \BL^{-1} \By = (\BL_x^{-1} \otimes \BL_f^{-1}) \By \in \realset{N}$, by using \eqref{eq:linSysKro}, we then have:
\begin{align}
	\BA = \BL_f^{-1} \BY (\BL_x^{-1})^\top =  ({\BL_x}^{-1} [{\BL_f}^{-1} {\BY}]^\top)^{\top} = ({\BL_x}^{-1} \BB^\top)^{\top},
	\label{eq:alphaKro}
\end{align}
where $\BB = {\BL_f}^{-1} {\BY}$. Hence, to construct the matrix $\BB \in \realset{R \times N}$, we need to solve the linear system $\BL_f \BB = \BY$, which can be efficiently computed since $\BL_f$ is a lower triangular matrix. Similarly, matrix $\BA \in \realset{R \times S}$ is obtained by solving the triangular-structured system ${\BL_x} \BA^\top = \BB^\top$. Finally, vector $\Ba$ results from vectorizing the matrix $\BA$.

Note that for computing \eqref{eq:quadraticTerm} using \eqref{eq:alphaKro}, intermediate steps require constructing and storing $R \times N$ and $R \times S$ matrices rather than the inverse of the $N \times N$ covariance matrix. In our application, this led to significant computational improvements since we considered large numbers of spatial design points $S$.

\subsection{Computation of the conditional mean and covariance functions}
\label{app:kronGPs:subapp:kronPred}
From \eqref{eq:condGPEqs}, we can note that the conditional mean function $\mu$ and the conditional covariance function $c$ also depend on the computation of $\BK$. As in Appendix \ref{app:kronGPs:subapp:kronLik}, here, we provide efficient computations of those quantities by exploiting the properties of the Kronecker product (see Appendix \ref{app:kronGPs:subapp:kron}).

We first focus on the computation of $\mu$. For the ease of the notation, we denote $\mu := \mu(\Bx_\ast, \BfF_\ast)$ and
$$\Bk := \Bk(\Bx_\ast, \BfF_\ast) = [k((\Bx_\ast,\BfF_\ast), (\Bx_1,\BfF_1)), \ldots, k((\Bx_\ast,\BfF_\ast),  (\Bx_N,\BfF_N))]^\top.$$
Then, following a similar procedure as the one used in Appendix \ref{app:kronGPs:subapp:kronLik} and using \eqref{eq:kron}, we can show that $\mu$ is given by
\begin{align*}
	\mu
	= {\Bk}^\top \BK^{-1} {\By}
	= {\Bk}^\top ({\BL_x}^\top \otimes {\BL_f}^\top)^{-1} {\Ba},
\end{align*}
with $\Ba = (\BL_x^{-1} \otimes \BL_f^{-1}) \By$. Now, since $\Bk = \Bk_x \otimes \Bk_f$ with $\Bk_x = [k_x(\Bx_\ast,\Bx_1), \ldots, k_x(\Bx_\ast,\Bx_S)]^\top$ and $\Bk_f = [k_f(\BfF_\ast,\BfF_1), \ldots, k_f(\BfF_\ast,\BfF_R)]^\top$, \eqref{eq:kron} yields:
\begin{align}
	\mu
	= ([{\BL_x}^{-1} {\Bk_x}]^\top  \otimes [{\BL_f}^{-1} {\Bk_f}]^\top) {\Ba}.
	\label{eq:muKronOpt}
\end{align}
As discussed in Appendix \ref{app:kronGPs:subapp:kronLik}, the computations $\Bb_x = {\BL_x}^{-1} {\Bk_x}$ and $\Bb_f = {\BL_f}^{-1} {\Bk_f}$ are efficient since $\BL_x$ and $\BL_f$ are lower triangular matrices. Then, \eqref{eq:muKronOpt} is obtained by applying \eqref{eq:linSysKro}.

Now, denote $c := c((\Bx_\ast, \BfF_\ast), (\Bx'_\ast, \BfF'_\ast))$, $k := k((\Bx_\ast,\BfF_\ast), (\Bx'_\ast,\BfF'_\ast))$ and $\Bk' := \Bk(\Bx'_\ast, \BfF'_\ast) = [k((\Bx_1,\BfF_1),(\Bx'_\ast,\BfF'_\ast)), \ldots, k((\Bx_N,\BfF_N),(\Bx'_\ast,\BfF'_\ast))]^\top$. Following a similar procedure as the one used for $\mu$, we have that $c$ is given by
\begin{align}
	c 
	= k - \Bk^\top \BK^{-1} \Bk'
	= {k} - (\Bb_x \otimes \Bb_f)^\top (\Bb'_x \otimes \Bb'_f)
	= {k} - (\Bb_x^\top \Bb'_x) (\Bb_f \Bb'_f),
\end{align}
with $\Bb_x = {\BL_x}^{-1} {\Bk_x}$, $\Bb_f = {\BL_f}^{-1} {\Bk_f}$, $\Bb'_x = {\BL_x}^{-1} {\Bk'_x}$, and $\Bb_f = {\BL_f}^{-1} {\Bk'_f}$.

\bibliographystyle{apa}  
\bibliography{references}

\end{document}